\documentclass[journal]{contents/vgtc}                     

\onlineid{1443}

\vgtccategory{Research}

\vgtcpapertype{application/design study}

\title{The Role of Visualization in LLM-Assisted Knowledge Graph Systems: Effects on User Trust, Exploration, and Workflows}

\author{%
  Harry Li,
  Gabriel Appleby, 
  Kenneth Alperin,
  Steven R Gomez, and
  Ashley Suh
}

\abstract{%
Knowledge graphs (KGs) are powerful data structures, but exploring them effectively remains difficult for even expert users. Large language models (LLMs) are increasingly used to address this gap, yet little is known empirically about how their usage with KGs shapes user trust, exploration strategies, or downstream decision-making---raising key design challenges for LLM-based KG visual analysis systems. To study these effects, we developed LinkQ, a KG exploration system that converts natural language questions into structured queries with an LLM. We collaborated with KG experts to design five visual mechanisms that help users assess the accuracy of both KG queries and LLM responses: an LLM-KG state diagram that illustrates which stage of the exploration pipeline LinkQ is in, a query editor displaying the generated query paired with an LLM explanation, an entity-relation ID table showing extracted KG entities and relations with semantic descriptions, a query structure graph that depicts the path traversed in the KG, and an interactive graph visualization of query results. From a qualitative evaluation with 14 practitioners, we found that users---even KG experts---tended to overtrust LinkQ's outputs due to its ``helpful'' visualizations, even when the LLM was incorrect. Users exhibited distinct workflows depending on their prior familiarity with KGs and LLMs, challenging the assumption that these systems are one-size-fits-all---despite often being designed as if they are. Our findings highlight the risks of false trust in LLM-assisted data analysis tools and the need for further investigation into the role of visualization as a mitigation technique. 
}


\teaser{
  \centering
  \includegraphics[width=.95\linewidth]{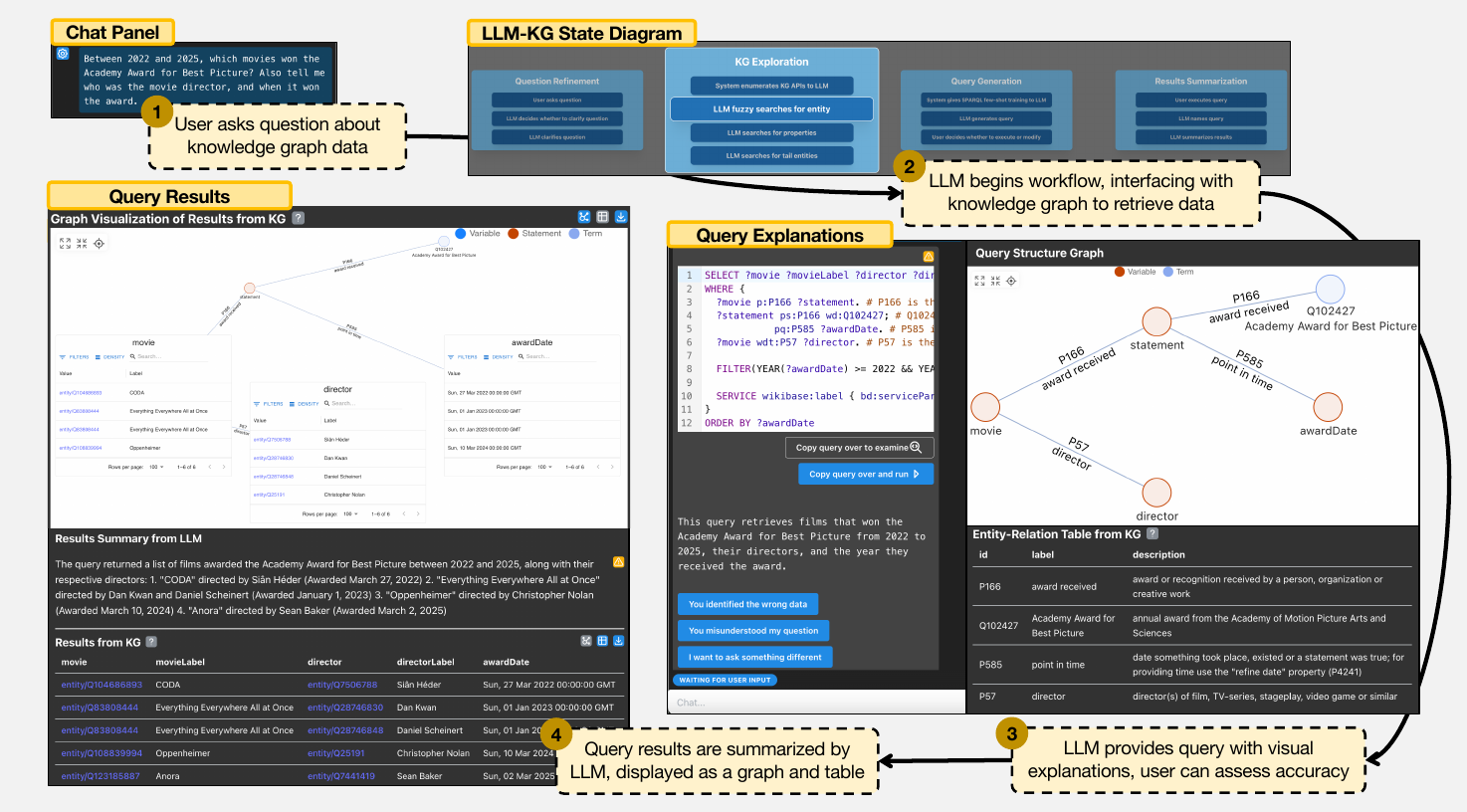}
  \caption{The \sys system, an LLM-assisted KG visual exploration system developed in collaboration with KG practitioners. Following the demonstrated workflow, we use \sys as a testbed for understanding the role of visualization in LLM-based KG exploration systems.
  }
  \label{fig:teaser}
}

\graphicspath{{figs/}{figures/}{pictures/}{images/}{./}} 

\usepackage{tabularx}
\usepackage{tabulary}
\usepackage{capt-of}

\usepackage{csquotes}

\usepackage{color, colortbl}

\usepackage{makecell}

\usepackage{booktabs,siunitx}

\usepackage{lipsum}

\usepackage{enumitem}

\usepackage{subfigure}

\usepackage{subcaption}
\usepackage{wrapfig}
\usepackage{xspace}
\usepackage{float}
\usepackage{graphicx}

\usepackage[normalem]{ulem}
\definecolor{mred}{rgb}{.80,.12,.30}
\definecolor{MRED}{rgb}{.80,.12,.30}
\definecolor{grey}{rgb}{0.5,0.5,0.5}
\definecolor{purple}{rgb}{.75,0,.85}
\definecolor{pistachio}{rgb}{0.58, 0.77, 0.45}
\definecolor{palesilver}{rgb}{0.9, 0.9, 0.9}

\newcommand{\sys}{LinkQ\xspace}

\newif\ifnotes
\notestrue

\let\origcite\cite
\renewcommand{\cite}[1]{\ifnotes\mbox{\origcite{#1}}\else \origcite{#1}\fi}

\begin{document}



\maketitle

\section{Introduction}

The widespread use of knowledge graphs (KGs) has enabled large-scale question-answering, recommendation systems, and visualization tools across industry and academia\cite{hogan2021knowledge}. 
Yet, their accessibility remains a persistent challenge due to their inherent complexity: knowledge graphs model real-world data as a collection of millions to billions of nodes, edges, and attributes. Because of this, knowledge graphs can be difficult to manage, retrieve data from, and explore effectively\cite{lissandrini2022knowledge, li2021kg4vis}.

While data visualization and interactive data management systems have become standard approaches to addressing these challenges, an emerging solution now includes the use of Large Language Models (LLMs). LLMs are able to update KG data, resolve entity ambiguities, and facilitate data exploration through query construction\cite{2024_unifying_llms_and_kgs}. However, we find that many LLM-assisted KG systems are introduced without studying their downstream impact on user workflows. Too often, humans are left completely out of the loop, with systems instead opting for multiple LLMs to handle the assessment and validation of KG data retrieval\cite{yang2023llm, arazzi2025augmented}. As a result, best practices for designing visualization-driven KG systems that leverage the strengths of LLMs---while mitigating their inherent limitations\cite{rawte2023survey}---remain underexplored.

In this paper, we present LinkQ: a KG exploration system designed to help users critically assess the accuracy and reliability of LLM-generated queries through several visual mechanisms. Unlike prior approaches that primarily seek to improve query accuracy from a technical standpoint, LinkQ was developed collaboratively with KG practitioners to explicitly keep end-users involved throughout the query-generation process. Consequently, by grounding the system's design in real-world practices, LinkQ serves as testbed for exploring how visualization shapes user interactions, trust, exploration strategies, and analytic processes in LLM-assisted KG exploration systems. 

We conducted a qualitative study with 14 practitioners experienced with both KGs and LLMs. Participants completed a combination of targeted question-answering tasks as well as open-ended tasks using two different KGs: the general-purpose Wikidata KG\cite{wikidata2024stats}, and the domain-specific BRON cybersecurity KG\cite{hemberg2020linking}. 
To best understand the role of visualization in LinkQ, our study intentionally included questions that we have found the LLM fails to retrieve the correct answer for---identified through a series of quantitative tests---as well as open-ended tasks that require more nuanced exploration strategies beyond simple data retrieval.

Although LinkQ was overwhelmingly praised for its \textit{``intuitive''} visualization designs, we found evidence of end-users \textbf{overtrusting} its outputs---even when those outputs were incorrect. 
This occurred when users found the query structure graph (which visualizes the traversal method for data extraction) aligned with their own mental model of how the KG should be queried, or when the LLM generated a results summary that seemed plausible. When the query output was incorrect---even blatantly incorrect---some users rationalized LinkQ's answer by externalizing circumstances that could justify its (false) output.

Users exhibited distinct workflows and exploration strategies while completing their tasks---based on their prior experience and skepticism toward LLMs---suggesting that LLM-assisted KG systems should not be designed as one-size-fits-all solutions. Those with less KG experience found little value in the query visualizations and preferred to only interface with LinkQ through its chat panel; in contrast, users with more KG experience diligently studied LinkQ's query graph and entity-relation visualizations to ensure that the query seemed accurate. Our more-skeptical users interacted deeply with LinkQ to double-check sources and citations for where the query outputs came from (e.g., linked citations to online articles). 

The remainder of the paper is structured as follows: We first overview related work on LLM-assisted KG systems and visualization techniques (Section~\ref{sec:related}). Next, we detail our practitioner-informed design process (Section~\ref{sec:goals}) that guided LinkQ's development (Section~\ref{sec:design}). We then present findings from our qualitative study (Section~\ref{sec:qualitative-user-evaluation}), discuss implications for visualization and system design (Section~\ref{sec:limitations}), and outline directions for future research (Section~\ref{sec:future}). Ultimately, this work contributes toward a deeper understanding of visualization's nuanced role in supporting LLM-driven KG exploration.

\section{Related Work}
\label{sec:related}

\subsection{Knowledge Graphs: Background \& Challenges}

Knowledge graphs (KGs) store complex, meaningful concepts as a network of nodes (\textit{entities}), edges (\textit{relations}), and properties (\textit{attributes}). A particular strength of KGs is their flexibility---any semantic relation can exist between two entities, and entities can possess any number of attributes. The structure and constraints defining these relationships form the graph's \textit{schema}. KGs are foundational for many large-scale question-answering systems\cite{ehrlinger2016towards, wei2020combining}, and are often the backbone to common language tasks---particularly when paired with LLMs\cite{alkhamissi2022review, petroni2019language, fewShotPaper}. See\cite{hogan2021knowledge, 2024_unifying_llms_and_kgs} for a set of diverse applications. 

Despite their advantages, the practical challenges of using KGs are well-documented\cite{li2024kgs, lissandrini2022knowledge, lissandrini2020graph}. Common issues include inconsistent data quality, difficulty maintaining relationships, visualizing large-scale KGs, and extracting relevant insights via querying. Querying KGs typically requires understanding multiple querying languages that differ across graph database platforms (e.g., Neo4J uses Cypher, while Stardog uses SPARQL\cite{hogan2021knowledge}). Although KGs are flexible, their query languages require precise pattern-matching, which often requires a deep understanding of the underlying schema\cite{li2024kgs}. Crafting correct queries is therefore tedious, preventing even expert KG users from acquiring relevant data for downstream analysis. 

Recent work has proposed using LLMs to alleviate these challenges\cite{li2024preliminaryroadmapllmsassistants, 2024_unifying_llms_and_kgs}. LinkQ builds upon these theoretical works by implementing question-answering over KGs via LLM query construction. 
For the purpose of addressing our research goal, LinkQ primarily supports querying RDF-based knowledge graphs\cite{RDF}; however, we demonstrate using \sys with other KG representations in our supplemental material.
For more details on the RDF data model, and how it differs from other KG representations, see\cite{w3corg}.

\subsection{Natural Language Interfaces \& LLMs}

Natural language interfaces (NLIs) have long supported users in various data analysis and question-answering tasks\cite{Aurisano2016Articulate2, Narechania2020NL4DVAT, mitra2022facilitating, huang2023flownl}, enabling exploration and visualization without requiring prior database or querying expertise\cite{li2014constructing}. Traditional NLIs suffer from difficulty in interpreting vague or ambiguous user tasks, which can be more easily deciphered with LLMs\cite{karanikolas2023large}. Consequently, LLMs are often used as interactive assistants for data analysis tasks; for example, for hypothesis testing\cite{chainForge}, for personalizing chatbot behaviors\cite{ha2024clochat}, for exploring prompt design spaces\cite{suh2024luminate}, for generating aethestically-pleasing images\cite{wang2024promptcharm}, and for authoring data-driven stories\cite{sultanum2023datatales}.
LLMs have also shown promise for visual analysis tasks\cite{basole2024generative}, such as detecting misleading visualizations\cite{alexander2024can, tian2024chartgpt}, identifying design flaws\cite{bendeck2024empirical}, and generating charts from user prompts\cite{tian2024chartgpt}. 

Within the KG domain, NLIs have historically been used to guide users in generating structured queries via natural language\cite{ferre2017sparklis, ngonga2013sorry, grafkin2016sparql} or through visualization\cite{vargas2019rdf}. The recent use of LLMs can improve KG exploration due to their ability to infer intent and reframe user questions into query representations. Some approaches work by fine-tuning\cite{zhang2023dissecting} LLMs to write accurate KG queries\cite{yang2023llm, rangel2024sparql}; however, these methods are typically not generalizable and can suffer poor performance on KGs the LLM was not fine-tuned on\cite{yang2023llm}. 

Beyond fine-tuned approaches, general-purpose LLMs are also used as `traversal agents' (for KG data retrieval), `reasoning agents' (for KG question-answering), and `construction agents' (for keeping the KG up-to-date with new data, or by resolving entities and deduplications)\cite{wen2024mindmap, knowledgeInjecitonToCounter, 2024_unifying_llms_and_kgs}. For a broad overview on the dual use of KGs and LLMs, see~\cite{petroni2019language, alkhamissi2022review}. In \sys, we use a general-purpose LLM (i.e. not fine-tuned) to support users in constructing KG queries for question-answering. Our design of \sys encourages iterative back-and-forth dialogue to refine user questions, and includes multiple visualizations to help assess the accuracy of both queries and LLM responses. Unlike traditional NLIs, \sys enables more nuanced interpretations of a user's question, inferring both the context of the question and the data domain being queried. Altogether, we leverage these features as a testbed for understanding users' workflows and exploration strategies when interacting with LLM-based KG visualization systems. 


\subsection{Effects on User Trust and Workflows}

The integration of LLMs into visualization tools introduces new risks---particularly in how users interpret, trust, and rely on LLM outputs, regardless of their correctness\cite{rawte2023survey}. 
Research in explainable AI has shown that LLMs can inadvertently lead to overreliance on AI\cite{buccinca2021trust}, resulting in overtrust for model recommendations (even when errors are present)\cite{he2025conversational, ha2024guided}, or can confuse users when LLM outputs clash with their mental models of the data\cite{crisan2024exploring}.

In evaluating LLM-driven systems, the primary research objective typically consists of understanding, \textit{``is the LLM sufficient at supporting users in its intended task,''} rather than investigating the more realistic scenario: \textit{``how are workflows, data exploration, and trust impacted when conducting analysis with an LLM assistant?"} 
This gap is particularly evident in KG exploration tools, where many suggest improved usability through LLM usage but lack empirical evidence demonstrating the downstream impacts on user behaviors\cite{arazzi2025augmented, 2024_unifying_llms_and_kgs, wen2024mindmap, fernandez2023large, agrawal-etal-2024-knowledge}. 

Studies in human-centered AI show that, when AI-generated outputs are paired with plausible-sounding justifications, users are more likely to accept incorrect information\cite{jacovi2023diagnosing, lipton2018mythos}. Our study findings of \sys align with these results, as users often trusted sensibly-structured query visualizations and LLM explanations, even when the results were incorrect. Research in fostering trust often encourages transparency through visualization design\cite{sultanum2024data, mckinley2025trustworthy, crouser2024building}; yet, we find that LinkQ's use of visualization can lead to overtrust in the system's outputs. Notably, this differs from other research that studies how visualization can intentionally or unintentionally deceive users~\cite{lee2021viral, lisnic2023misleading}; LinkQ was designed to aid interpretation, but produced unintended overtrust.

Moreover, we find that integrating an LLM into KG exploration can impact users' analysis workflows, shaping their strategies for interacting with data, interpreting visualizations, and validating information. Understanding these adjustments is crucial for designing LLM-assisted KG systems, we discuss these implications further in Section~\ref{sec:limitations}.
\begin{figure*}[t]
    \centering
    \includegraphics[width=.9\linewidth]{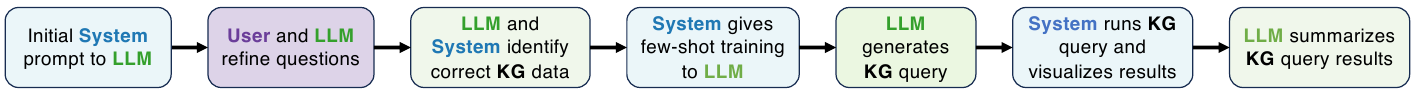}
    \caption{A high-level overview of LinkQ's\cite{li2024linkq} prompting protocol, described in Section~\ref{sec:prompting-protocol}. For this pipeline, the \textbf{LLM}, \textbf{System}, and \textbf{User} have precise responsibilities for completing the question-to-query translation.
    }
    \label{fig:pipeline}
\end{figure*}

\section{Design Requirements}
\label{sec:goals}

\begin{figure*}[p]
    \centering
    \includegraphics[width=\textwidth, height=\textheight, keepaspectratio]
    {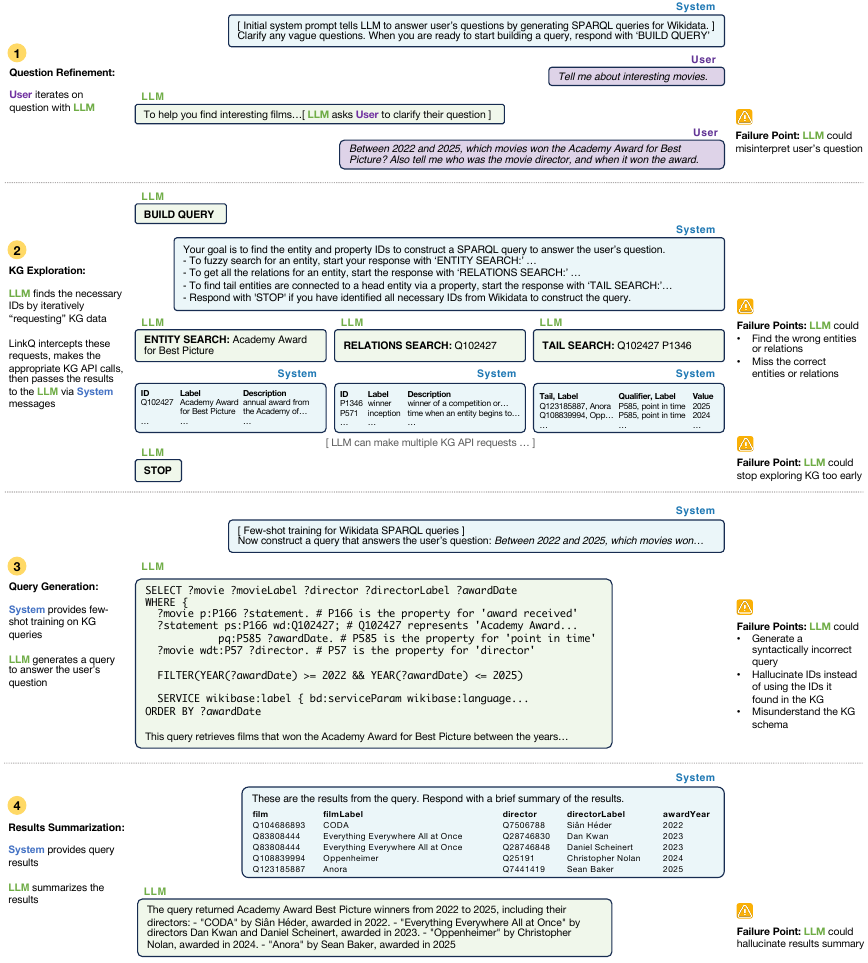}
    \caption{An example of the LinkQ prompting protocol described in Section \ref{sec:prompting-protocol}, based off the scenario in Figure \ref{fig:teaser}. Initially, the LLM iterates with the User to clarify their question. Then the LLM searches for ground-truth data in the KG. Once the LLM has found IDs it needs, it generates a query, then summarizes and LinkQ visualizes the results. Each stage of the workflow is depicted in a state diagram at the top of LinkQ, and users can choose to view this full conversation history in the user interface. On the right side of the figure, we note possible LLM failure points in the process.
    }
\label{fig:full_conversation_history_directors}
\end{figure*}





Our goal in developing \sys was to explicitly bring human judgment back into the loop of LLM-assisted KG exploration, allowing us to study how different visualization components influence user trust, workflows, and downstream analysis. We followed an iterative, user-centered design process to identify system requirements that would be both representative of KG practitioners' real-world needs and effective for studying visualization’s impact on user interactions with LLMs.

Early in our design process, we built upon prior research in KG query builders\cite{ell2015spartiqulation, grafkin2016sparql, ferre2017sparklis}, natural language question-answering interfaces\cite{srinivasan2021snowy, sultanum2023datatales, mitra2022facilitating, huang2023flownl}, and recent KG-LLM roadmaps\cite{lissandrini2022knowledge, 2024_unifying_llms_and_kgs, li2024preliminaryroadmapllmsassistants}. We aimed to address limitations of previous rigid, rule-based query builders\cite{vargas2019rdf} and incorporate chain-of-thought strategies for more explainable LLMs\cite{wen2024mindmap, feng2023knowledgeSolver}. Grounding \sys in existing literature ensured our evaluation would reflect open research challenges.

Following best practices in design study methodologies\cite{harte2017human, selmdmair2012dsm}, we collaborated with four data scientists who regularly work with KGs. We first surveyed their preferred types of questions, interactions, and visualization modalities that an LLM-based KG system should support. We then iterated on system sketches and prototypes through a series of three feedback sessions to ensure the visualizations and interactions made sense when paired with their workflows. A complete overview of this process---including initial sketches and detailed feedback---is available in our supplemental material.

Collectively, we defined the following design goals for \sys:

\begin{enumerate}[topsep=2pt, partopsep=0pt,itemsep=2pt,parsep=2pt,label=\textbf{G\arabic*}]
    \item \label{goal:refine} \textbf{Enable iterative, human-in-the-loop refinement of natural language questions into precise KG queries.} When a user's question is exploratory in nature, the LLM should help refine any vague or ambiguous qualifiers with data in the KG. A targeted question should ultimately result in a well-formed KG query.
    
    \item \label{goal:no-hallucinate} \textbf{Minimize and explicitly surface LLM errors (e.g., \textit{hallucinations}).} Generating KG queries with an LLM may result in erroneous data IDs or false query results. The system should include measures to reduce the frequency of hallucinations, while explicitly highlighting when and where they may occur---empowering users to identify and correct them on their own.
    
    \item \label{goal:preview} \textbf{Provide query previews and contextual explanations to support user verification.} Given the computational cost and risk of incorrect queries for downstream analysis, users must be provided with transparent previews and intuitive explanations of LLM outputs. These visual and textual explanations should provide users with sufficient context to judge query correctness and validity, helping further mitigate LLM-produced errors.

    \item \label{goal:multimodal} \textbf{Present query results in multimodal formats to facilitate verification and exploration.} Users engage in KG question-answering for a variety of reasons; therefore, \sys should offer multimodal outputs: text summaries for interpretation, tabular views for systematic verification, and graphical visualizations for exploratory tasks. Presenting multiple modalities supports users in cross-validating query results to reinforce their trust and understanding.
\end{enumerate}

\section{LinkQ}
\label{sec:design}
\sys employs a chained prompting approach\cite{wei2022chain}, overviewed in Figure \ref{fig:pipeline} and demonstrated in-depth in Figure~\ref{fig:full_conversation_history_directors}, which we refer to as LinkQ's ``protocol.'' The exact prompts we use in LinkQ are included in our supplemental and our code is open-source at \textit{(redacted)}.
An important distinction to make in \sys is the different responsibilities for the \textbf{LLM}, \textbf{System}, and \textbf{User}. The LLM is the language model that helps process and interpret text from the User and System; the backend System relays messages, prompts, and API calls between the LLM and KG; and the User converses with the LLM without having to directly interact with the KG or System.

\subsection{KG and LLM Integrations}
\label{sec:kg-integration}
\noindent 
\textbf{LLM Integration:}
LinkQ can integrate with any LLM that is exposed through an OpenAI-compatible API server. Users can use OpenAI GPT models, or alternatively host their own LLM through an open-source tool (like vLLM\cite{vLLM}), which creates OpenAI-compatible API servers for public foundational models (such as Llama3\cite{grattafiori2024llama3herdmodels}). 

\smallbreak 
\noindent 
\textbf{KG Integration:}
Since KGs vary widely in their structure\cite{hogan2021knowledge}, we developed LinkQ to perform question-answering over RDF-based KGs\cite{RDF} via the KG query language SPARQL. 
For generalizability, we have modularly structured LinkQ to make it easy to change the KG data source, which involves updating:

\begin{enumerate}[topsep=2pt, partopsep=0pt,itemsep=0pt,parsep=0pt]
  \item Initial system prompt that describes the KG data and schema.
  \item API calls to the KG to explore the data and find the right IDs.
  \item Few-shot training\cite{fewShotPaper} on a respective KG querying language.
\end{enumerate}

The open-source version of LinkQ is built-in with the \textbf{Wikidata} KG as it is well-maintained, generalizable, and has a robust API service. We also integrated LinkQ with the \textbf{BRON} cybersecurity KG\cite{hemberg2020linking} for the purpose of our qualitative evaluation (Section~\ref{sec:qualitative-user-evaluation}), as our study participants have expertise in the cyber domain. In our supplemental material, we demonstrate using the three above steps to integrate \sys with non-RDF KGs (such as Neo4j and the Cypher querying language\cite{miller2013graph}).

\subsection{LinkQ Querying Protocol}
\label{sec:prompting-protocol}
\sys's prompting protocol (Figure \ref{fig:full_conversation_history_directors}) guides the LLM through multiple iterative steps to query the KG based on the user's questions. 

\vspace{2pt}
\noindent 
\textbf{Question Refinement}: The LLM is instructed in its initial system prompt to clarify a user's ambiguous or open-ended question through a back-and-forth dialogue (\ref{goal:refine}). The KG schema is provided to the LLM so it can suggest relevant data to the user during question refinement. Once the LLM decides the user's question is well-formed enough, it begins the KG exploration process to generate a query.

\vspace{2pt}
\noindent 
\textbf{KG Exploration}: {During query construction, the LLM is tasked first with finding all the KG entity and relation IDs it needs for the query. The LLM and System work together to fuzzy search for entities, find relevant relations, and traverse the KG to identify all the correct IDs. Whenever the LLM wants to access data in the KG, LinkQ calls the appropriate KG API and passes the KG graph structure and IDs to the LLM via System messages (\ref{goal:no-hallucinate}).
}

\vspace{2pt}
\noindent 
\textbf{Query Generation}: {When the LLM has identified all the necessary IDs, the System provides the LLM with few-shot training\cite{fewShotPaper} to write an accurate SPARQL query. By having the LLM generate a query instead of directly answering the user's question, LinkQ both mitigates LLM hallucinations and can retrieve up-to-date data (\ref{goal:no-hallucinate}).}

\vspace{2pt}
\noindent 
\textbf{Results Summarization}: {When the user executes a query, LinkQ visually displays the query results in both tabular and graphical format and also provides an LLM-generated summary of the results (\ref{goal:multimodal}).}


\subsection{LinkQ UI Components and Visualizations}

We developed the LinkQ UI (Figure~\ref{fig:teaser}) as a natural language interface that helps users answer questions, generate queries with the LLM, inspect whether the generated queries are accurate, and view the results. In particular, the visualizations in LinkQ intend to illuminate failure points that the LLM can suffer from (see the right panel of Figure~\ref{fig:full_conversation_history_directors}).

\smallbreak 
\noindent
\textbf{Chat Panel:} %
Other than as a communication device, the LLM suggests data from the KG and provides explanations about its queries inside the Chat Panel. Buttons are shown to the user and can be selected if: (1) the LLM identified the wrong data; (2) the LLM misunderstood the question; or (3) the user wants to ask a different question. Each button produces a templated prompt that can be edited by the user.

\smallbreak 
\noindent
\textbf{LLM-KG State Diagram:} %
The top panel of LinkQ's NLI is a state diagram visualization that provides users with a transparent mental model of how the system works. The state diagram displays four primary prompting protocol states: (1) Question Refinement, (2) KG Exploration, (3) Query Generation, and (4) Results Summarization. Each state also shows a detailed breakdown of sub-states. As the user interacts with the tool, LinkQ highlights the precise state the system is in, e.g., ``\textit{KG Exploration $\rightarrow$ LLM fuzzy searches for entity.}'' The left panel in Figure~\ref{fig:full_conversation_history_directors} describes each respective state in detail. 

\smallbreak 
\noindent
\textbf{Query Editor with Explanation:} %
%
The Query Editor is a standard code interface with keyword highlighting, helping users assess the syntax of the query.
When the LLM generates a query, it also provides an explanation of (1) what the query does via in-line comments, displayed in the editor; and (2) how the query addresses the user's question, displayed in the chat panel. This helps users who are unfamiliar with the querying language understand how the query works (\ref{goal:preview}).
\smallbreak 
\noindent
\textbf{Entity-Relation Table:} %
%
The Entity-Relation Table extracts the entity and relation IDs in the query and provides their associated human-readable labels and descriptions. For example in Wikidata, the entity ``\textit{Q102427}'' has the label	``\textit{Academy Award for Best Picture,}'' with the description, ``\textit{annual award from the Academy of Motion Picture Arts and Sciences.}'' This feature helps users quickly assess whether the LLM has correctly found (or hallucinated) the relevant IDs.
\smallbreak 
\noindent
\textbf{Query Structure Graph:} %
The Query Graph shows a visual preview of a syntactically valid query in graphical form (\ref{goal:preview}). This graphic allows the user to better understand the relationships within their queries and the potential KG schema.
The Query Graph specifically supports SELECT queries that consist of basic graph patterns, similar to the visual query builder RDF Explorer\cite{vargas2019rdf}.
All heads and tails (i.e. entities) are represented as graphical nodes, and all relations are represented as graphical edges. Nodes and edges are labeled by their corresponding text within the query, and a human-readable label is provided if it can be extracted from the KG. Resolved variables are colored blue, and unresolved variables are colored orange. 
Even if the user is unfamiliar with the query language, the Query Graph can be used to assess the accuracy of a query by helping the user understand the variables that the query will resolve (i.e. pattern match on), as well as which relationships exist between data in the query. This helps the user mitigate the potential for executing generated queries that are syntactically correct, but semantically incorrect.

\begin{figure}[H]
    \centering
    \includegraphics[width=0.49\textwidth]{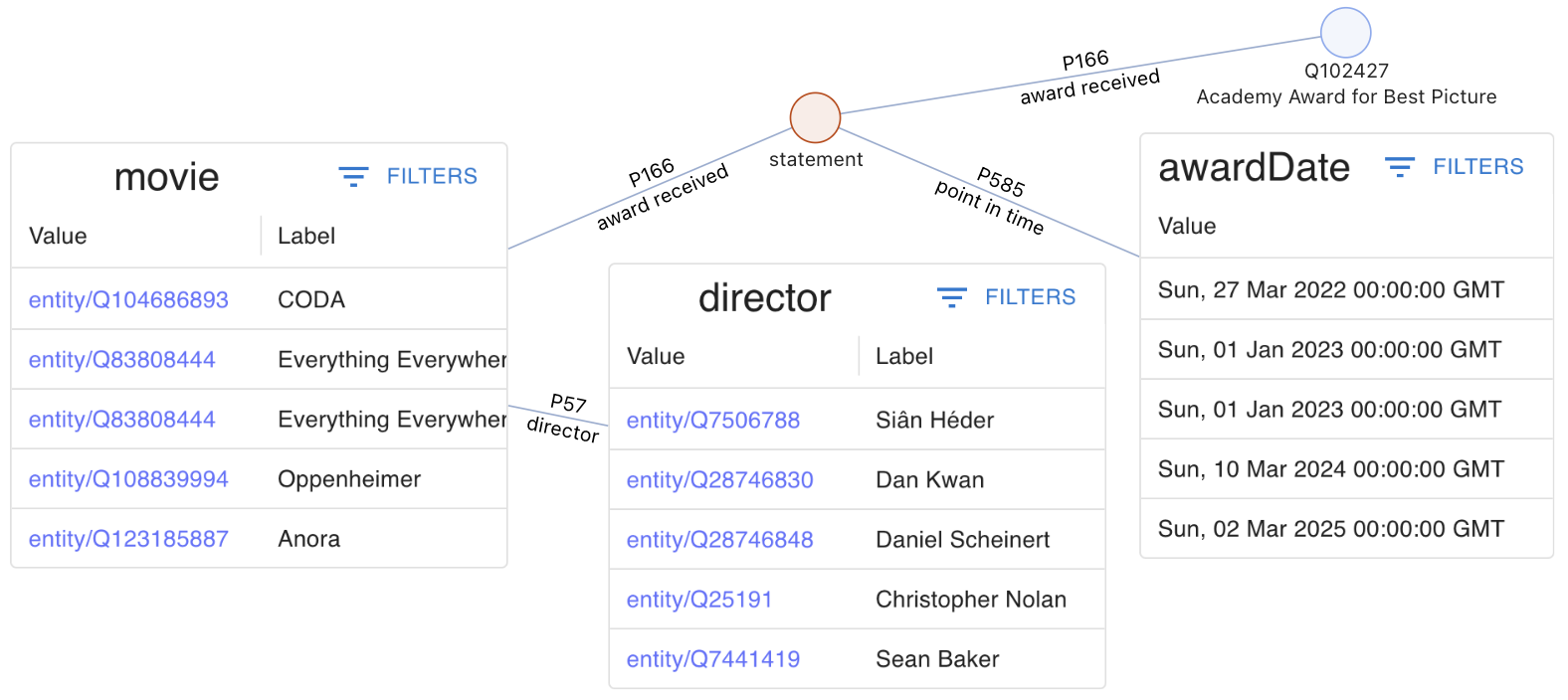}
    \caption{Example of LinkQ's Results Graph Visualization, where nodes (represented as a table of entities) and edges (the relations connecting entities) are extracted from the KG.}
    \label{fig:graph-results}
\end{figure}
\smallbreak 
\noindent
\textbf{Results Graph Visualization:} %
The Results Graph (Figure~\ref{fig:graph-results}) extends the metaphor of the Query Structure Graph, replacing each variable node with a tables of results, while retaining the existing edges. For example, the unresolved variable \texttt{?director} becomes a node in the results graph called `directors,' where all directors retrieved from the query appear as a table. 
This makes it easy to follow how the query results were extracted from the KG, while mitigating clutter in the graph\cite{li2024kgs}.
Each column of the tabular results replaces the variable node and becomes its own table embedded within the graph.
Hovering over any row within that table scrolls every other embedded table to their matching row from the result set (i.e. a multi-coordinated view). 

\smallbreak 
\noindent
\raisebox{-0.1em}{\includegraphics[height=1em]{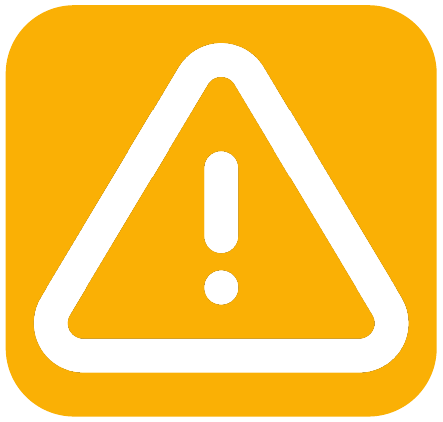}}\hspace{0.1em} \textbf{Hallucination Warnings:} %
Wherever the LLM generates content within LinkQ, we add a caution icon with a pop-up that explains the risk of hallucination. This helps users distinguish between ground-truth results from the KG versus LLM-generated text with potential hallucinations. See the right panel in Figure~\ref{fig:full_conversation_history_directors} for these `failure points.' 

\subsection{Quantitative Evaluation}
\label{sec:quantitative-mintaka-evaluation}

We conducted a small quantitative evaluation of LinkQ's prompting protocol (described in Section~\ref{sec:prompting-protocol}) to verify it performs better than a standalone LLM for KG querying. We tested both LinkQ and GPT-4 (our comparative baseline) on a total of 120 questions from the Mintaka\cite{sen2022mintaka} Wikidata knowledge graph question bank. Each question belonged to one of five categories: Multi-hop, Comparative, Yes/No, Generic, and Intersection. 
A complete write-up of our evaluation is available as supplemental material.

\begin{table}[h!]
\centering
\small
\renewcommand{\arraystretch}{1}
\centering
\resizebox{.9\linewidth}{!}{%
\renewcommand\theadalign{bt}
\renewcommand\theadfont{\bfseries}
\renewcommand\theadgape{\Gape[4pt]}
\renewcommand\cellgape{\Gape[4pt]}
\setlength{\tabcolsep}{8pt}

\small
\sffamily
\begin{tabular}{lll}
\toprule
\textbf{Question Type} & \textbf{LinkQ Accuracy} & \textbf{GPT-4 Accuracy}          
\\
\midrule 
Comparative  & 91.7\% & 20.8\%  \\
Yes/No       & 87.5\% & 54.2\%  \\
Generic      & 79.2\% & 33.3\%  \\
Multi-Hop    & 75.0\% & 16.7\%  \\ 
Intersection & 54.2\% & 12.5\%  \\ 
\bottomrule
\end{tabular}

}
\caption{Results of our quantitative evaluation comparing LinkQ to GPT-4 (our baseline) on query accuracy for the Mintaka KBQA dataset\cite{sen2022mintaka}.}
\label{tab:quant-eval}
\end{table}

Overall, LinkQ outperformed GPT-4 in accuracy for every question type tested, indicating that LinkQ's prompting strategy can indeed improve the objective correctness of an LLM translating natural language questions into KG queries. Table~\ref{tab:quant-eval} shows the results. LinkQ performed by far the best on Comparative questions (LinkQ: 91.7\%, GPT-4: 20.8\%), and performed the worst on Intersection questions (LinkQ: 54.2\%, GPT-4: 12.5\%).


Our quantitative results suggest that systems like LinkQ can perform well on writing KG queries that look for data existence, conduct fact-checking, and retrieve data from one--three hops away. This said, it is clear that prompt-engineered LLMs will not produce perfect results for KG querying---at least, without significant fine-tuning---thus requiring a human-in-the-loop approach. We discuss further in Section~\ref{sec:future}.

\begin{figure*}[h]
    \centering
    \includegraphics[width=.85\linewidth]{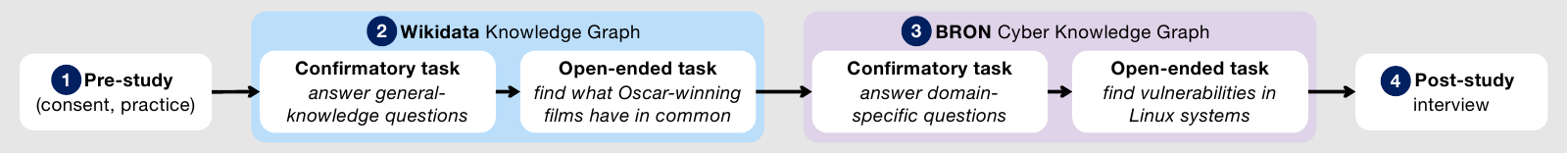}
    \caption{Process of our qualitative user study (Section~\ref{sec:qualitative-user-evaluation}). Participants completed a series of tasks with LinkQ using two different KGs, beginning with Wikidata (a general-purpose KG) and ending with BRON (a domain-specific cybersecurity KG). 
    }
    \label{fig:eval-workflow}
\end{figure*}

\section{Qualitative Study}
\label{sec:qualitative-user-evaluation}

\subsection{Study Design}

We conducted a qualitative, think-aloud study\cite{north2006toward} of the LinkQ system, largely basing its design on recent studies for LLM data analysis systems\cite{wang2024promptcharm, chainForge, ha2024clochat, sultanum2023datatales}. Our goal was to understand: \textit{``How is a user's analysis workflow, exploration strategies, and overall trust impacted when using an LLM-assisted KG visualization system for question-answering?''}
Figure~\ref{fig:eval-workflow} shows an overview of our study design, and a detailed write-up is included in our supplemental material.

\begin{itemize}[leftmargin=0pt,topsep=4pt, partopsep=0pt,itemsep=2pt,parsep=2pt]
    \item[] {\textbf{Protocol:} Participants were recruited from a professional network of practitioners who currently or have previously worked with KGs and LLMs. Each evaluation lasted between 1--1.5 hours and were conducted either in-person or virtually using a video conferencing tool. All participants consented to the study prior to its start. Two authors conducted a qualitative, think-aloud study that consisted of questions that had pre-determined correct or incorrect answers, as well as open-ended questions that had no correct or incorrect answer. Participants were told that they would not be graded by correctness but were still asked to describe their confidence in the answers generated by LinkQ to help us elicit their trust in different visualization components. Participants were told upfront that LinkQ would not always get the correct or expected answer for their questions, this was demonstrated in a live walkthrough of LinkQ answering the question, \textit{``Tell me something interesting about the Harry Potter series''}, which we verified beforehand returns unexpected results from the Wikidata KG.  Each component of the LinkQ was described and shown to participants before beginning their usage of the system. Each participant used the tool during the study---either on a laptop in-person, or through screen-control virtually. Finally, participants were asked a series of follow-up questions to gather their feedback on the tool. The resultant qualitative data was analyzed with a focus on emergent themes\cite{curry2009qualitative} related to our original research goals.}

    \item[] {\textbf{Participants:} We followed a saturation-based approach\cite{guest2006many} in which participants were recruited until no new themes emerged during data collection. Our evaluation consisted of 14 participants, with 3/14 having been a part of the initial stages of developing LinkQ (Section~\ref{sec:goals}). Participants have a BS (3/14), an MS (7/14) or a PhD (4/14). All participants were required to have previous experience in working with KGs and LLMs, but still had varying experience levels. Participants ranked their experience for both on a Likert scale of 1 (none) to 5 (extreme), shown in Table \ref{tab:participants}. For the remainder of our study, we classify participants as `KG experts' when they have a self-reported experience level of 4--5 (8/14), `KG non-experts' as those with an experience level of 2 (4/14) and those who are `familiar but not experts' as those with an experience level of 3 (2/14). Most participants identified as `LLM experts' with a self-reported score of 4--5 (11/14) with a remaining (3/14) being `familiar but not experts' with a self reported score of 3. All of the participants in our study work in AI, ML, and data science with a focus in the cybersecurity domain. We discuss the implications of our participants' education levels as well as their KG and LLM expertise further in Section~\ref{sec:study-limitations}.}

    \item[] {\textbf{Knowledge Graphs: } Participants completed an equal number of tasks with two different KGs: the Wikidata KG\cite{wikidata2024stats} and the BRON cybersecurity KG~\cite{hemberg2020linking}. We selected Wikidata KG because of its domain-agnostic, general-purpose use that all participants were familiar with. In contrast, we chose the BRON KG because of its domain-specificity in cybersecurity. While all of our participants are knowledgeable in cyber, and familiar with the topics for each of the tasks and questions in the study, only 3/14 had used BRON previously.} 

    \item[] {\textbf{Targeted questions:} For the Wikidata use case, we selected questions from the Mintaka question bank\cite{sen2022mintaka} that (1) LinkQ tended to get correct, and (2) LinkQ tended to get incorrect (as per our quantitative evaluation in Section~\ref{sec:quantitative-mintaka-evaluation}). For the BRON use case, we selected questions that are sensible for a realistic usage scenario of BRON based on its original publication\cite{hemberg2020linking}. Participants were given an equal amount of time to answer Wikidata and BRON questions; not all participants answered all questions in the time allotted.}

    \item[] {\textbf{Open-ended questions:} For both KGs, we presented participants with an open-ended question to answer with LinkQ. Both questions described scenarios in which they needed to explore the particular KG to generate a report for their stakeholder. For Wikidata, the report was to describe the commonalities in recent films that have won an Academy Award for Best Picture. For BRON, the report was to describe weaknesses and vulnerabilities associated with Linux systems.}

\end{itemize}

\subsection{Visualizations Usage}
Participants visually communicated with and interpreted LinkQ's responses differently depending on their prior KG expertise. The LLM-KG State Diagram fostered the most trust in the LLM's capabilities, while the Query Structure Graph was the most helpful in assessing the LLM's reliability for those with at least some KG querying experience. 

\begin{itemize}[leftmargin=0pt,topsep=4pt, partopsep=0pt,itemsep=2pt,parsep=2pt]

    \item[] \textbf{LLM-KG State Diagram pulls back the curtain on LLMs:} The state diagram was by far the favorite visual component, with participants often suggesting it \textit{``shed light''} and added trust into the LLM's workflow: \textit{``I really like the state diagram while it's running because it's giving so much insight into what's going on\ldots One, it's a lot more engaging than waiting for the query, and two, it's helping me understand what's actually happening in the process.''} Participants noted that they would like \textit{every} LLM-assisted tool to feature a similar state diagram; however, some remarked that the details could be trimmed over time: \textit{``Once you get familiar with the system, then an abbreviated version of the state diagram might be enough\ldots you won't need to see every single state.''}

    \item[] \textbf{Entity-Relation Table provided context to both KG experts and non-experts:} Although not flashy, the entity-relation table provided helpful context for the data contained within the LLM's generated query: \textit{``The descriptions from the entity ID table gives me more confidence in the query.''} It was especially helpful for KG non-experts, who told us that \textit{``anything like this helps us verify the LLM's answer,'}' without having to diligently study the query structure itself. 

    \item[] \textbf{Query Editor and Query Structure Graph were highly used by KG experts, but the Chat Panel was preferred by non-experts:} Participants familiar with KGs, even if not experts at writing queries, spent a majority of the task time inspecting queries as well as the query structure graphs: \textit{``I'm looking at the graph here, and it looks like it's capturing what I wanted it to. I'm not super familiar with SPARQL so I'm not super equipped to read this, but it makes sense for the most part.''} However, for participants with no KG querying experience, they were often confused by the query structure: \textit{``I don't understand what these variables mean\ldots the relationship nodes are not always immediately interpretable.''} One participant told us that they would rather not have the query graph visible at all: \textit{``I would be fine with not seeing anything there. I'd be fine with the LLM just showing me the answer and giving me a source for its information. I don't want to have to look at the query. As long as there's some way for me to click on a source to verify the answer [within the chat panel], I'd be fine with that.''} 

    \item[] \textbf{Graph Visualization of Results was never used, regardless of KG expertise:} Overall, no participants chose to use the graph visualization of results as their primary view; instead, all participants opted for viewing the results in tabular format (the KG Results Table). Some remarked that the graph visualization would be helpful as graphics inserted into PowerPoints or visual presentations (i.e. as eye candy\cite{li2024kgs}). One participant noted that the graph visualization would be more useful if it showed additional information from the KG---that was not directly queried about---to contextualize the query results, e.g., as an exploratory tool rather than a confirmatory one. 

\end{itemize}

\subsection{Workflows \& Exploration Strategies}
\label{sec:eval-workflows}
We observed participants adopting distinct strategies for assessing LinkQ's accuracy as well as for answering their analysis questions. 

\begin{itemize}[leftmargin=0pt,topsep=4pt, partopsep=0pt,itemsep=2pt,parsep=2pt]

    \item[] \textbf{Deliberate question-answering over open-ended exploration:} 13/14 participants told us that---while LinkQ seemed more effective for targeted question-answering---they \textit{preferred} using it for open-ended tasks: \textit{``I enjoyed using LinkQ for open-ended questions because it gave me the freedom to problem-solve exactly the way I want. I liked that I could speak to it naturally and it was able to intuit my thought process.''} Despite this preference, we observed participants were very deliberate in how they approached open-ended tasks with LinkQ, focusing on a series of direct questions rather than extensively exploring or iteratively refining open questions with the LLM. We discuss the implications of this finding further in Section~\ref{sec:limited-exploration}.
    
    \item[] \textbf{Debugging queries with the LLM and error widgets:} 8/14 participants collaborated with the LLM to refine their questions and correct its generated queries, typically using the LLM summary of results to explain what should be different: \textit{``You returned a list of films and its directors, but I actually wanted information about...'' } Often participants used the prompt widgets (buttons in the Chat Panel, seen in Figure~\ref{fig:teaser}) as a starting point to correct the LLM, which was highly praised as a convenient way to help debug queries.

    \item[] \textbf{Validating outputs through sources and linked references:} 7/14 participants confirmed answers by exploring the sources provided in LinkQ's KG Results Table (e.g., clicking on the link to the entity page to view its KG entry directly). For example, when checking the birthplace of the 2019 Wimbledon winner, participants clicked on the entity link for LinkQ's answer (``Novak Djokovic''), navigated to Djokovic's Wimbledon statistics page to verify he won the 2019 championship, then finally checked his birthplace. 
    Participants expressed lower confidence in LinkQ's responses when they could not verify them through this linked references process.
    
    \item[] \textbf{Cross-checking answers through related entities and attributes in the KG:} 5/14 participants verified answers by systematically cross-checking related information from the KG. For example, when asking which member of the Black Eyed Peas was in the movie Poseidon, they used LinkQ's answer (``Fergie'') to confirm: (1) if Fergie was indeed a member of the Black Eyed Peas, (2) the full cast of Poseidon, and whether Fergie was in the list, and (3) the inception date of both the Black Eyed Peas and the movie Poseidon.

    \item[] \textbf{Implicitly trusting LinkQ without further verification:} 4/14 participants accepted LinkQ's answer at face value without attempting further validation: \textit{``I don’t really know whether it's right or not, but it says it is. So that must be the answer.''} This behavior occurred more frequently with the BRON KG, as participants felt their unfamiliarity limited their ability to assess the query or LLM outputs on their own.
\end{itemize}

\subsection{Differences Based on Expertise}
We found that, depending on participants' prior experience with KGs and LLMs, their approach to interacting with the system greatly varied. 

\begin{itemize}[leftmargin=0pt,topsep=4pt, partopsep=0pt,itemsep=2pt,parsep=2pt]

    \item[] \textbf{Cleverly phrasing questions, based on their experience, to get a well-written query:} Participants with more KG experience intentionally phrased their questions so that the queries would be simple to generate for LinkQ. For example, when tasked with multi-hop questions, some participants broke them up into multiple parts: \textit{``I think it makes more sense to phrase this [multi-hop question] as two parts: Who won the men's Wimbledon in 2019, and where were they born?''} In contrast, when participants had less KG experience but more LLM experience, they tended to provide additional context about their question to LinkQ beyond what the task had originally written. For example, when asking about cast members for the movie Poseidon, one participant explained to LinkQ: \textit{``I don't want to know the movie character names. I want the names of the individuals who played those characters.''}
    
    \item[] \textbf{KG experts cautiously inspected query structures:} Participants with deeper KG experience would spend a greater amount of time sanity-checking the KG query generated by LinkQ before executing it. If the query structure seemed plausible (i.e.~matched their own mental model of what the query \textit{should} look like), these participants generally tended to believe the answer was correct---even if at times, it was not. For example, after inspecting and confirming the query structure ``looked good,'' one participant told us he was certain the answer would be correct, even before executing the query. Interestingly, these participants also dedicated more time to inspecting queries for the BRON KG, suggesting greater caution when (1) they were less familiar with the KG, and (2) had greater domain expertise over the tasks.

    \item[] \textbf{KG non-experts never inspected queries before executing them:} Participants who had little KG experience almost never looked at the KG queries before running them: \textit{``I don’t know SPARQL so I don't know how to tell whether it's correctly written.''} One participant told us that, because of his in-depth experience using LLMs, he generally tended to trust and have faith in the LLM's ability to complete its task (i.e.~he assumed LinkQ would get the answer correct regardless). However, participants with little KG experience but high LLM skepticism would still spend time trying to verify LinkQ's outputs, although they felt less comfortable doing this through query inspection. In these cases, participants used other components of the interface (e.g., the entity-relation table or the LLM's explanation of the query) to get more context for the query, but not necessarily to verify its structural correctness as KG experts did. 
\end{itemize}






\subsection{Limitations \& Areas of Improvements}
Based on our observations as well as participant feedback, we identified several limitations and opportunities for improving systems like LinkQ.

\begin{itemize}[leftmargin=0pt,topsep=4pt, partopsep=0pt,itemsep=2pt,parsep=2pt]

    \item[] \textbf{Trust and explainability:} Participants sometimes implicitly trusted LinkQ's responses when unsure about the correct answer themselves---occasionally even rationalizing incorrect answers as valid (more on this in Section~\ref{sec:vis-role}). This issue was exacerbated by insufficient explanations from LinkQ when queries returned empty results or encountered SPARQL limitations, leaving participants confused about whether the problem came from the KG or the LLM itself. Participants recommended that LinkQ could be more transparent in its responses---to assure correctness, or illuminate failure points---by breaking complex queries into discrete, thoroughly-explained reasoning steps.

    \item[] \textbf{Contextualization and semantic richness:} LinkQ’s effectiveness was at times limited by the semantic descriptiveness of the underlying KG. Participants wanted LinkQ to provide additional contextual information and richer semantic details---even when not explicitly asked. For example, when queries for BRON returned entities by identifier alone (e.g., a software vulnerability labeled \textit{``cve:CVE:2000''}), participants wanted LinkQ to ``know'' that it should include descriptions for vaguely-named entities. Similarly, query results occasionally lacked sufficient contextual metadata, e.g., when querying box office earnings of films, LinkQ did not indicate whether the earnings represented global or regional markets, making it difficult for users to interpret the results. Participants remarked that external integration with non-KG resources for easier validation, including additional citations, external sources, or online references, could help contextualize results.

    \item[] \textbf{Query refinement and LLM flexibility:} There were a few times that participants expressed frustration with the LLM, remarking that its behavior was unintelligent. This included its inability to build upon previously generated queries during follow-up questions---instead, the LLM would construct new queries from scratch, even when incremental changes would have been more efficient. There were also times when the LLM opted to answer's the user question (via query construction) even though it had knowledge that the data did not exist based on the KG's schema. Instead, it would be more helpful if the LLM would respond \textit{``that data does not exist, but here is something similar...''} 

    \item[] \textbf{Performance and user control integration:} Query runtime was identified as an issue for some questions, especially complex questions that required more than two hops. Participants suggested that LinkQ could be prompted to optimize queries by integrating `performance hints.' Similarly, participants requested better control mechanisms for LinkQ; for example, the ability to filter, modify, or reset query parameters.
  
\end{itemize}

\section{Design Implications}
\label{sec:limitations}

\subsection{Fostering Necessary Trust in LLM Outputs}
Users' skepticism for LLMs can vary: some extend much more faith in their capabilities (\textit{``The fact that I know about n-gram tags, like what perplexity does, it gives me confidence in LLMs''}), while some believe Generative AI is altogether unusable for nonmenial tasks (\textit{``When I do a Google Search, I just skip over the AI-generated summary because it's always wrong.''}) We had several participants in our study remark that they are \textit{``predisposed to be cautious of LLMs,''} describing themselves as \textit{``someone whose default is not to be super trusting of LLMs.''} 

Our LLM-skeptical participants overwhelmingly found LinkQ facilitated trust in the LLM's outputs, as they knew its responses came from ground truth data in the KG: \textit{``I felt like I was seeing the steps the LLM went through, when I asked it for who won Wimbledon 2019, it went and found what entry in the KG represented Wimbledon 2019, then built a query. Yup, that looks like it's going to tell me what I wanted it to.''} One participant specifically compared it to a more trustworthy ChatGPT: \textit{``I would never trust this answer from ChatGPT, I know LinkQ's answer isn't coming from its training. The fact that its answer isn't constructed from a sentence based on a probability of words definitely gives me more trust.''} Even our participants with little KG experience told us they would prefer to use LinkQ (specifically, an LLM-based querying agent) for question-answering tasks.  

While LinkQ demonstrated that trust in LLMs can be improved by answering user questions via querying ground-truth data, there are further opportunities to increase the trust and usability of LLMs. 
Currently, \sys relies on the LLM to tell the user if something might have gone wrong with its generated query, but future systems can investigate a more visualization-centric way to interpret empty results, as it is not always the case that there is no data relevant to the query. One possible solution could be to show semantically similar data from the KG directly (e.g., using an embedding-based approach) when the results are empty or limited. Similarly, during initial question-refinement, future systems could incorporate a rules-based or grammar-based approach to help visually guide exploration, so long as users do not become frustrated with rigid translations\cite{srinivasan2021snowy, setlur2016eviza}. 

\subsection{Visualization's Role in Explaining (Bad) LLM Behavior}
\label{sec:vis-role}
All participants were told upfront that LinkQ might get answers incorrect, and our interface contains multiple caution icons that indicate where outputs can contain errors or be falsely produced (e.g., because of an incorrectly written query or incorrect entity resolution -- see Figure~\ref{fig:full_conversation_history_directors}). That said, we found participants often \textit{overtrusted} the LLM's answer, often citing LinkQ's visualizations paired with LLM explanations as reasons why they believed the (false) answer was correct.

At times during our study, LinkQ would retrieve an answer that participants initially thought could be wrong, but proceeded to externalize reasons why LinkQ was probably right: \textit{``It says Gladiator won best picture in 2023, but isn't that movie old? I know a second one came out recently, so maybe that's when it won?''} Even when performing question-answering over domain-specific data that participants were familiar with (i.e. with the BRON KG), participants would still justify incorrect outputs from LinkQ that they were uncertain about. For example, when retrieving vulnerabilities associated with Google Chrome, LinkQ (incorrectly) returned no results. Our participant told us, \textit{``I think that makes sense, since Google Chrome probably has a lot of support, it probably doesn't have any vulnerabilities.''} 

Based on the think-aloud nature of our study, in addition to participant feedback, we believe that the added visual transparency within LinkQ's natural language interface may have resulted in overconfidence in the LLM's outputs. The biggest culprit---while also being participants' favorite visualization---is likely the LLM-KG State Diagram. The insights into what the LLM is doing at every step tended to sway participants in believing the LLM was always accurately performing its tasks. For those with high KG experience, the Query Structure Graph tended to sway them into believing the query outputs were correct, as the query \textit{``looked reasonable,''} and \textit{``should get the right answer.''}

Our findings align with emergent studies that show how an LLM's tendency to ``smooth over'' inconsistencies can result in overreliance on AI\cite{buccinca2021trust, he2025conversational}. Transparently designed visualizations, while helpful for revealing LLM reasoning, may inadvertently exacerbate overtrust by reducing users' scrutiny of outputs, as suggested (unintentionally) by one participant: \textit{``I really liked that the interface was flexible enough that someone could accept the information at face value.''}  
Similar to current research in explainable AI, it is likely that visualization design spaces will need to shift towards \textit{challenging} users' assumptions around LLM outputs---rather than purely clarifying system behavior---especially in this new era of LLM-assisted data analysis tools. 

Potential strategies could include visualizations that explicitly highlight where uncertainty exists in the system workflow (perhaps in a more pronounced manner than LinkQ's caution icons), as well as proactively suggesting contradictory or alternative interpretations via in-line visualizations. One participant suggested to us that alternative queries, subgraphs of the KG query space, and an embedding view of similar entities or relations would help add transparency into the LLM's query accuracy: \textit{``It would be helpful to see what facts the LLM chose from when creating the query, like what `Poseidon' did the it choose from given the whole set of available `Poseidon' entities?''}

Moreover, when the output of these LLM systems is numerical (e.g., for more traditional data analysis tasks), participants suggested \textit{``when I'm asking about numbers or counts, I'd like to see raw data, raw numbers, a summary for the counts across different groups or entities to see how it all adds up,''} suggesting that the visualization design space should map appropriate visual explanations from the LLM's output type. When an LLM's output is not numerical, participants told us: \textit{``I want a more in-depth explanation from the LLM for how the query addresses my question. That's the best way for me to know if the query results are correct.''} Due to the complexity of KGs, we believe there will not always be \textit{precise} visualizations that can universally illustrate whether an LLM's response is correct. In these cases, system designers might need to compartmentalize the LLM's functionality into small units and explain them one-by-one to the user, visually highlighting where mistakes could be encountered.

Altogether, our findings suggest that LLM-assisted data analysis systems are \textbf{not one-size-fits all}, despite often being designed as if they are. The visualizations depicted in these systems will need to vary depending on: (1) the LLM's output type (e.g., visually explaining the steps an LLM took to produce an analysis, versus showing visual alternatives for an LLM's recommendations); (2) its use case (e.g., for open-ended exploration versus targeted question-answering); and (3) its user base (e.g., highly skeptical LLM users versus trusting ones). Emergent work in trustworthy visualization design\cite{mckinley2025trustworthy, ha2024guided} can help developers understand when trust might be falsely gained through paired visualization and LLM design. 

\subsection{LLMs May Limit Visual Data Exploration}
\label{sec:limited-exploration}
As briefly discussed in Section~\ref{sec:eval-workflows}, we found that participants approached exploratory analysis with LinkQ in an unexpected manner. We observed that participants rarely attempted to have the LLM first answer the overarching question (e.g., ``what do Oscar-winning films have in common?'') and instead opted for first thinking aloud what types of traits films might have in common, then asking LinkQ about those traits one-by-one (e.g., \textit{``what are the genres of films that have won an Oscar in the last five years?''}). We believe this implicit behavior aligns with the behavior stemming from belief elicitation\cite{koonchanok2021data, koonchanok2023visual}, in which users tend to conduct very deliberate, focused analyses when asked to first externalize their beliefs about the data.  

Although a user's analysis is more deliberate, these studies also find that the data is overall less explored during belief elicitation sessions. We observed a similar finding during LinkQ's evaluation, in which users tended to not begin their exploration for open-ended tasks in a top-down manner (e.g., not starting with finding all possible attributes of award-winning films, then narrowing down their search) but rather focusing on a bottom-up approach. It is possible that conducting analysis with an LLM assistant produces less coverage of the data, with users potentially focusing on directed questions that they feel more comfortable with the LLM answering. Future work can consider studying this concept directly, e.g., comparing data coverage with and without an LLM.

Finally, an important consideration is that---for now---LLMs are incapable of perfect responses, in LinkQ's case: perfectly-written queries. This can either be due to a lack of context about a knowledge graph, because of a misunderstanding of the data's structure contained in the KG, or due to misinterpreting the user's semantic intent. 
When the LLM writes an incorrect query, what happens when results are empty or limited? The user might believe that there are simply no results related to their query, or that perhaps there is no existing data in the KG for their question---we encountered this phenomena in our qualitative study. Does this mean a user will believe their analysis is concluded?  Moreover, what happens when the LLM retrieves an incorrect answer but wrote a correct query? These open questions are interesting avenues for future qualitative (even wizard-of-oz\cite{kelley1984iterative}) studies to better inform the effects of LLM-assisted visual exploratory systems. 

\section{Discussion}
\label{sec:future}

\subsection{Preliminary Guidelines for Designing LLM-KG Systems}

Our qualitative findings raise open questions about visualization’s role in the era of LLMs. Here, we offer preliminary guidelines distilled from our evaluation of an LLM-assisted KG question-answering system:

\begin{enumerate}[label=\textbf{G\arabic*},leftmargin=17pt,topsep=4pt, partopsep=0pt,itemsep=2pt,parsep=2pt]

\item \textbf{Support human-in-the-loop query construction.} 
Generating a successful KG query from a natural language question should be considered a multi-step, human-in-the-loop process---as KG queries are complex in nature and will rarely result in perfect LLM performance. Our Chat Panel, Entity-Relation Table, and Query Structure Graph were all consistently utilized to help users refine their queries alongside the LLM.

\item \textbf{Align visualizations to user expertise while explicitly counteracting overtrust.}
Visual support during query generation should reflect user expertise: a KG expert might only want to see a subset of the KG being queried, while a non-expert might want to see tables of semantic data descriptors for added context. Importantly, visualizations should provide views that challenge outputs and alert users of potential errors or gaps in the query---not just for transparency's sake but to mitigate overtrust. Best practices in trustworthy visualization\cite{sultanum2024data, crouser2024building} and avoiding unintentional deception\cite{lee2021viral, lisnic2023misleading} can inform this design space.


\item \textbf{Streamline query refinement and task specification.} 
Presenting all possible human-readable labels, data descriptions, and metadata from the KG can help users correct an LLM's generated query. Features such as \textit{prompt widgets} (interactive buttons for inserting customizable prompts) and interactive highlighting for pinpointing incorrect query segments can simplify refinement. Finally, allowing users to explicitly state their question-answering goals upfront (e.g., in system instructions) can help guide the LLM towards retrieving the most relevant data.

\item \textbf{Deploy unique querying and explanation strategies for different question types.}
Not all question types are equally easy for an LLM to translate into queries. Future systems should consider categorizing questions by complexity and designing unique strategies for (1) tasking the LLM with translating them into queries, and (2) explaining the process to the user. 
When the LLM is unsuccessful, visual interfaces should include step-by-step collaborative methods for the user to assist the LLM in fixing the query. 

\end{enumerate}

\subsection{Extending our Qualitative Study}
\label{sec:study-limitations}
Our qualitative study involved 14 participants belonging to different departments within two organizations. Our participants are highly educated individuals, highly experienced in data science, and displayed inquisitive behaviors when interacting with LinkQ.  All participants had at least \textit{some} to very high KG and LLM experience, suggesting that our findings may not generalize to ``every day users'' who are unfamiliar with KGs, LLMs, or trustworthy design. Consequently, future studies in LLM-assisted visual analysis systems can consider comparing to the design implications presented in Section~\ref{sec:limitations}. These studies may extend the research questions in this paper to examine how trust is impacted in users who are \textit{not} skeptical of AI, and consequently may not make the effort to check linked resources and citations in LLM-based systems. 

The set-up of our qualitative study was also limited by task and time constraints. Asking practitioners to take more than an hour of their work day to participate in an evaluation is in and of itself challenging\cite{rogers2012hci}. In an ideal world, we could investigate our research questions in a longitudinal study, observing how participants' usage and trust in LinkQ changes over time. Future studies can also consider extending our study to include tasks beyond targeted and open-ended question-answering, or, more rigorously testing the quantitative differences in using an LLM-assisted KG system versus a non-LLM assisted KG system. Moreover, while we found workflow and exploration strategies were dependent on user expertise, it is possible that with enough time getting comfortable with LinkQ, these strategies could also change. 

With that said, the LinkQ system is not an inherently complex system. LinkQ's overarching goal is to answer users' questions about KG data by having an LLM generate and explain KG queries. In the future, features that further support open-ended exploration, sensemaking, custom chart creation, and so on will continue to change the role of visualization in LLM-assisted analysis systems. These features can be considered for building upon the current state of LinkQ.

\subsection{Extending LinkQ}
While the focus of this work was studying the effects of visualization in LLM-assisted KG systems, there are interesting avenues of future work for improving LinkQ (as well as systems like LinkQ).

\begin{itemize}[leftmargin=0pt,topsep=4pt, partopsep=0pt,itemsep=2pt,parsep=2pt]

    \item[] \textbf{Retrieval Augmented Generation:} LinkQ implements a chained prompting protocol to improve an LLM's ability to generate correct KG queries for question-answering. In the future, a text embedding-based approach to inject relevant documents into the prompts (i.e. Retrieval Augmented Generation (RAG)\cite{rag}) could be integrated to improve accuracy and output transparency. For example, rather than LinkQ's current method of fuzzy searching for entities in the KG by entity name, LinkQ can be extended to find entities by comparing a search string with text embeddings of entity names and descriptions.

    \item[] \textbf{Performance and cost:} Future LLM-driven text-to-query systems should not wholly rely on an LLM as much as \sys does, although we (among others\cite{hong2024nextgenerationdatabaseinterfacessurvey}) have shown that powerful models like GPT-4 do very well at text parsing and entity extraction. While a trade-off in performance is possible, cheaper models can alleviate the burden of costly OpenAI API calls and environmental risks\cite{rillig2023risks}. To improve the overall performance of the \sys protocol, a tool like ChainForge\cite{chainForge} could be used to optimize the prompts for the LLM. Additionally, other KG APIs could be added to the protocol that could lead to better query generation, for example, to find paths between two entities. These improvements could also potentially address our study participants' concerns about query execution time. 

    \item[] \textbf{Integrating \sys with non-KG databases:} While our focus has been supporting users in KG exploration, another promising direction is exploring how the \sys protocol can be applied to other kinds of structured data and querying languages, such as relational databases and SQL. The learning outcomes from these experiments can inform extensions of \sys with relational databases; moreover, the insights gained from these experiments can inform the broader design space of using LLMs as query generators, given a user's analysis goals and semantic intent.

\end{itemize}
\section{Conclusion}
We present \sys, an open-source, human-in-the-loop system for KG question-answering with an LLM. \sys supports users in asking natural language questions, either open-ended or targeted, which are iteratively refined with an LLM until they can be translated into well-formed KG queries. In collaboration with KG practitioners, we designed LinkQ's interface to include visualizations that provide transparency into the LLM's workflow, helping users examine the accuracy of the LLM's generated query. We conducted a qualitative study of LinkQ, exploring how visualization impacts user workflows, exploration strategies, and downstream analysis. Our participants appreciated that LinkQ helped them understand the inner workings of the LLM during their tasks; however, some users tended to overtrust the LLM's outputs due to LinkQ's ``intuitive'' visualization design. Users exhibited distinct workflows depending on their familiarity with KGs and skepticism for LLMs, suggesting LLM-KG systems should not be designed as one-size-fits-all. To conclude, we discuss design guidelines and future improvements for LinkQ based on the findings of our study.

\acknowledgments{
DISTRIBUTION STATEMENT A. Approved for public release. Distribution is unlimited.
This material is based upon work supported by the Combatant Commands under Air Force Contract No. FA8702-15-D-0001. Any opinions, findings, conclusions or recommendations expressed in this material are those of the author(s) and do not necessarily reflect the views of the Combatant Commands. © 2025 Massachusetts Institute of Technology. Delivered to the U.S. Government with Unlimited Rights, as defined in DFARS Part 252.227-7013 or 7014 (Feb 2014). Notwithstanding any copyright notice, U.S. Government rights in this work are defined by DFARS 252.227-7013 or DFARS 252.227-7014 as detailed above. Use of this work other than as specifically authorized by the U.S. Government may violate any copyrights that exist in this work.

\noindent This work was authored in part by the National Renewable Energy Laboratory (NREL), operated by Alliance for Sustainable Energy, LLC, for the U.S. Department of Energy (DOE) under Contract No. DE-AC36-08GO28308. The views expressed in the article do not necessarily represent the views of the DOE or the U.S. Government. The U.S. Government retains and the publisher, by accepting the article for publication, acknowledges that the U.S. Government retains a nonexclusive, paid-up, irrevocable, worldwide license to publish or reproduce the published form of this work, or allow others to do so, for U.S. Government purposes.
}

\appendix 

\section*{Appendix}

\begin{table}[h!]
\centering
\small
\renewcommand{\arraystretch}{1}
\centering
\resizebox{.8\linewidth}{!}{%
\renewcommand\theadalign{bt}
\renewcommand\theadfont{\bfseries}
\renewcommand\theadgape{\Gape[4pt]}
\renewcommand\cellgape{\Gape[4pt]}
\setlength{\tabcolsep}{8pt}

\small
\sffamily
\begin{tabular}{lll}
\toprule
\textbf{Experience} & \textbf{with KGs} & \textbf{with LLMs}          
\\
\midrule 
1 (None) & 0 / 14 & 0 / 14  \\
2 (Slight)     & 4 / 14 & 0 / 14  \\
3 (Some)       & 2 / 14 & 3 / 14  \\
4 (Moderate)   & 3 / 14 & 0 / 14  \\ 
5 (Extreme)    & 5 / 14 & 11 / 14  \\ 
\bottomrule
\end{tabular}

}
\caption{A breakdown of our participants' experience with KGs and LLMs from our qualitative study in Section~\ref{sec:qualitative-user-evaluation}. Experience was self-reported on a Likert scale of 1-5, where each level was described in terms of how often or thoroughly did they develop, train, interact with, or queried KGs and LLMs in their workplace.}
\label{tab:participants}
\end{table}

\bibliographystyle{abbrv-doi-hyperref}

\bibliography{main}

\begin{thebibliography}{10}

\bibitem{RDF}
{R}esource {D}escription {F}ramework ({R}{D}{F}) {M}odel and {S}yntax {S}pecification --- w3.org.
\newblock \url{https://www.w3.org/TR/PR-rdf-syntax/Overview.html}.
\newblock [Accessed 29-07-2024].

\bibitem{agrawal-etal-2024-knowledge}
G.~Agrawal, T.~Kumarage, Z.~Alghamdi, and H.~Liu.
\newblock Can knowledge graphs reduce hallucinations in {LLM}s? : A survey.
\newblock In K.~Duh, H.~Gomez, and S.~Bethard, eds., {\em Proc. NAACL: Hum. Lang. Tech. (Vol. 1: Long Papers)}, pp. 3947--3960. Association for Computational Linguistics, Mexico City, Mexico, June 2024. \href{https://doi.org/10.18653/v1/2024.naacl-long.219}
{doi: {{%
10\hspace{.1pt}\discretionary{.}{%
}{.}\hspace{.4pt}18653\discretionary{/}{%
}{/}v1\discretionary{/}{%
}{/}2024\hspace{.1pt}\discretionary{.}{%
}{.}\hspace{.4pt}naacl\discretionary{%
}{-}{-}long\hspace{.1pt}\discretionary{.}{%
}{.}\hspace{.4pt}219}}}


\bibitem{alexander2024can}
J.~Alexander, P.~Nanda, K.-C. Yang, and A.~Sarvghad.
\newblock Can gpt-4 models detect misleading visualizations?
\newblock In {\em 2024 IEEE Visualization and Visual Analytics (VIS)}, pp. 106--110. IEEE, 2024.

\bibitem{alkhamissi2022review}
B.~AlKhamissi, M.~Li, A.~Celikyilmaz, M.~Diab, and M.~Ghazvininejad.
\newblock A review on language models as knowledge bases.
\newblock {\em arXiv preprint arXiv:2204.06031}, 2022. \href{https://doi.org/10.48550/arXiv.2204.06031}
{doi: {{%
10\hspace{.1pt}\discretionary{.}{%
}{.}\hspace{.4pt}48550\discretionary{/}{%
}{/}arXiv\hspace{.1pt}\discretionary{.}{%
}{.}\hspace{.4pt}2204\hspace{.1pt}\discretionary{.}{%
}{.}\hspace{.4pt}06031}}}


\bibitem{chainForge}
I.~Arawjo, C.~Swoopes, P.~Vaithilingam, M.~Wattenberg, and E.~L. Glassman.
\newblock Chainforge: A visual toolkit for prompt engineering and llm hypothesis testing.
\newblock In {\em Proceedings of the CHI Conference on Human Factors in Computing Systems}, CHI '24,  article no. 304,  18 pages. Association for Computing Machinery, New York, NY, USA, 2024. \href{https://doi.org/10.1145/3613904.3642016}
{doi: {{%
10\hspace{.1pt}\discretionary{.}{%
}{.}\hspace{.4pt}1145\discretionary{/}{%
}{/}3613904\hspace{.1pt}\discretionary{.}{%
}{.}\hspace{.4pt}3642016}}}


\bibitem{arazzi2025augmented}
M.~Arazzi, D.~Ligari, S.~Nicolazzo, and A.~Nocera.
\newblock Augmented knowledge graph querying leveraging llms.
\newblock {\em arXiv preprint arXiv:2502.01298}, 2025.

\bibitem{Aurisano2016Articulate2}
J.~Aurisano, A.~Kumar, A.~Gonzales, J.~Leigh, B.~DiEugenio, and A.~Johnson.
\newblock Articulate 2 : Toward a conversational interface for visual data exploration.
\newblock In {\em Proc. VIS}, 2016.

\bibitem{basole2024generative}
R.~C. Basole and T.~Major.
\newblock Generative ai for visualization: Opportunities and challenges.
\newblock {\em IEEE Computer Graphics and Applications}, 44(2):55--64, 2024.

\bibitem{bendeck2024empirical}
A.~Bendeck and J.~Stasko.
\newblock An empirical evaluation of the gpt-4 multimodal language model on visualization literacy tasks.
\newblock {\em IEEE Transactions on Visualization and Computer Graphics}, 2024.

\bibitem{fewShotPaper}
T.~Brown, B.~Mann, N.~Ryder, M.~Subbiah, J.~D. Kaplan, P.~Dhariwal, A.~Neelakantan, P.~Shyam, G.~Sastry, A.~Askell, S.~Agarwal, A.~Herbert-Voss, G.~Krueger, T.~Henighan, R.~Child, A.~Ramesh, D.~Ziegler, J.~Wu, C.~Winter, C.~Hesse, M.~Chen, E.~Sigler, M.~Litwin, S.~Gray, B.~Chess, J.~Clark, C.~Berner, S.~McCandlish, A.~Radford, I.~Sutskever, and D.~Amodei.
\newblock Language models are few-shot learners.
\newblock In H.~Larochelle, M.~Ranzato, R.~Hadsell, M.~Balcan, and H.~Lin, eds., {\em Advances in Neural Information Processing Systems}, vol.~33, pp. 1877--1901. Curran Associates, Inc., 2020.

\bibitem{buccinca2021trust}
Z.~Bu{\c{c}}inca, M.~B. Malaya, and K.~Z. Gajos.
\newblock To trust or to think: cognitive forcing functions can reduce overreliance on ai in ai-assisted decision-making.
\newblock {\em Proceedings of the ACM on Human-computer Interaction}, 5(CSCW1):1--21, 2021.

\bibitem{crisan2024exploring}
A.~Crisan, N.~Butters, and Zoe.
\newblock Exploring subjective notions of explainability through counterfactual visualization of sentiment analysis.
\newblock In {\em 2024 IEEE Evaluation and Beyond-Methodological Approaches for Visualization (BELIV)}, pp. 15--24. IEEE, 2024.

\bibitem{crouser2024building}
R.~J. Crouser, S.~Matoussi, L.~Kung, S.~Pandey, O.~G. McKinley, and A.~Ottley.
\newblock Building and eroding: Exogenous and endogenous factors that influence subjective trust in visualization.
\newblock In {\em 2024 IEEE Visualization and Visual Analytics (VIS)}, pp. 306--310. IEEE, 2024.

\bibitem{curry2009qualitative}
L.~A. Curry, I.~M. Nembhard, and E.~H. Bradley.
\newblock Qualitative and mixed methods provide unique contributions to outcomes research.
\newblock {\em Circulation}, 119(10):1442--1452, 2009.

\bibitem{ehrlinger2016towards}
L.~Ehrlinger and W.~W{\"o}{\ss}.
\newblock Towards a definition of knowledge graphs.
\newblock {\em Proc. ESWC Posters and Demos Track}, 48(1-4):2, 2016.

\bibitem{ell2015spartiqulation}
B.~Ell, D.~Vrande{\v{c}}i{\'c}, and E.~Simperl.
\newblock Spartiqulation: Verbalizing sparql queries.
\newblock In {\em Proc. ESWC}, pp. 117--131. Springer, 2013.

\bibitem{grattafiori2024llama3herdmodels}
A.~G. et. al.
\newblock The llama 3 herd of models, 2024. \href{https://doi.org/10.48550/arXiv.2407.21783}
{doi: {{%
10\hspace{.1pt}\discretionary{.}{%
}{.}\hspace{.4pt}48550\discretionary{/}{%
}{/}arXiv\hspace{.1pt}\discretionary{.}{%
}{.}\hspace{.4pt}2407\hspace{.1pt}\discretionary{.}{%
}{.}\hspace{.4pt}21783}}}


\bibitem{feng2023knowledgeSolver}
C.~Feng, X.~Zhang, and Z.~Fei.
\newblock Knowledge solver: Teaching llms to search for domain knowledge from knowledge graphs, 2023.

\bibitem{fernandez2023large}
R.~C. Fernandez, A.~J. Elmore, M.~J. Franklin, S.~Krishnan, and C.~Tan.
\newblock How large language models will disrupt data management.
\newblock {\em Proc. VLDB}, 16(11):3302--3309, 2023.

\bibitem{ferre2017sparklis}
S.~Ferr{\'e}.
\newblock Sparklis: An expressive query builder for sparql endpoints with guidance in natural language.
\newblock {\em Semant. Web}, 8(3):405--418, 2017. \href{https://doi.org/10.3233/SW-150208}
{doi: {{%
10\hspace{.1pt}\discretionary{.}{%
}{.}\hspace{.4pt}3233\discretionary{/}{%
}{/}SW\discretionary{%
}{-}{-}150208}}}


\bibitem{grafkin2016sparql}
P.~Grafkin, M.~Mironov, M.~Fellmann, B.~Lantow, K.~Sandkuhl, and A.~V. Smirnov.
\newblock Sparql query builders: Overview and comparison.
\newblock In {\em BIR Workshops}, pp. 255--274, 2016.

\bibitem{guest2006many}
G.~Guest, A.~Bunce, and L.~Johnson.
\newblock How many interviews are enough? an experiment with data saturation and variability.
\newblock {\em Field methods}, 18(1):59--82, 2006.

\bibitem{ha2024clochat}
J.~Ha, H.~Jeon, D.~Han, J.~Seo, and C.~Oh.
\newblock Clochat: Understanding how people customize, interact, and experience personas in large language models.
\newblock In {\em Proceedings of the CHI Conference on Human Factors in Computing Systems}, pp. 1--24, 2024.

\bibitem{ha2024guided}
S.~Ha, S.~Monadjemi, and A.~Ottley.
\newblock Guided by ai: Navigating trust, bias, and data exploration in ai-guided visual analytics.
\newblock In {\em Computer Graphics Forum}, vol.~43, p. e15108. Wiley Online Library, 2024.

\bibitem{harte2017human}
R.~Harte, L.~Glynn, A.~Rodr{\'\i}guez-Molinero, P.~M. Baker, T.~Scharf, L.~R. Quinlan, G.~{\'O}Laighin, et~al.
\newblock A human-centered design methodology to enhance the usability, human factors, and user experience of connected health systems: a three-phase methodology.
\newblock {\em JMIR human factors}, 4(1):e5443, 2017.

\bibitem{he2025conversational}
G.~He, N.~Aishwarya, and U.~Gadiraju.
\newblock Is conversational xai all you need? human-ai decision making with a conversational xai assistant.
\newblock {\em arXiv preprint arXiv:2501.17546}, 2025.

\bibitem{hemberg2020linking}
E.~Hemberg, J.~Kelly, M.~Shlapentokh-Rothman, B.~Reinstadler, K.~Xu, N.~Rutar, and U.-M. O'Reilly.
\newblock Linking threat tactics, techniques, and patterns with defensive weaknesses, vulnerabilities and affected platform configurations for cyber hunting.
\newblock {\em arXiv preprint arXiv:2010.00533}, 2020. \href{https://doi.org/10.48550/arXiv.2010.00533}
{doi: {{%
10\hspace{.1pt}\discretionary{.}{%
}{.}\hspace{.4pt}48550\discretionary{/}{%
}{/}arXiv\hspace{.1pt}\discretionary{.}{%
}{.}\hspace{.4pt}2010\hspace{.1pt}\discretionary{.}{%
}{.}\hspace{.4pt}00533}}}


\bibitem{hogan2021knowledge}
A.~Hogan, E.~Blomqvist, M.~Cochez, C.~D’amato, G.~D. Melo, C.~Gutierrez, S.~Kirrane, J.~E.~L. Gayo, R.~Navigli, S.~Neumaier, A.-C.~N. Ngomo, A.~Polleres, S.~M. Rashid, A.~Rula, L.~Schmelzeisen, J.~Sequeda, S.~Staab, and A.~Zimmermann.
\newblock Knowledge graphs.
\newblock {\em ACM Comput. Surv.}, 54(4),  article no. 71,  37 pages, 2021. \href{https://doi.org/10.1145/3447772}
{doi: {{%
10\hspace{.1pt}\discretionary{.}{%
}{.}\hspace{.4pt}1145\discretionary{/}{%
}{/}3447772}}}


\bibitem{hong2024nextgenerationdatabaseinterfacessurvey}
Z.~Hong, Z.~Yuan, Q.~Zhang, H.~Chen, J.~Dong, F.~Huang, and X.~Huang.
\newblock Next-generation database interfaces: A survey of llm-based text-to-sql, 2024.

\bibitem{huang2023flownl}
J.~Huang, Y.~Xi, J.~Hu, and J.~Tao.
\newblock Flownl: Asking the flow data in natural languages.
\newblock {\em IEEE Trans. Vis. Comput. Graph.}, 29(1):1200--1210, 2023. \href{https://doi.org/10.1109/TVCG.2022.3209453}
{doi: {{%
10\hspace{.1pt}\discretionary{.}{%
}{.}\hspace{.4pt}1109\discretionary{/}{%
}{/}TVCG\hspace{.1pt}\discretionary{.}{%
}{.}\hspace{.4pt}2022\hspace{.1pt}\discretionary{.}{%
}{.}\hspace{.4pt}3209453}}}


\bibitem{jacovi2023diagnosing}
A.~Jacovi, J.~Bastings, S.~Gehrmann, Y.~Goldberg, and K.~Filippova.
\newblock Diagnosing ai explanation methods with folk concepts of behavior.
\newblock {\em Journal of Artificial Intelligence Research}, 78:459--489, 2023.

\bibitem{karanikolas2023large}
N.~Karanikolas, E.~Manga, N.~Samaridi, E.~Tousidou, and M.~Vassilakopoulos.
\newblock Large language models versus natural language understanding and generation.
\newblock In {\em In Proc. PCI}, pp. 278--290, 2023.

\bibitem{kelley1984iterative}
J.~F. Kelley.
\newblock An iterative design methodology for user-friendly natural language office information applications.
\newblock {\em ACM Transactions on Information Systems (TOIS)}, 2(1):26--41, 1984.

\bibitem{koonchanok2021data}
R.~Koonchanok, P.~Baser, A.~Sikharam, N.~K. Raveendranath, and K.~Reda.
\newblock Data prophecy: Exploring the effects of belief elicitation in visual analytics.
\newblock In {\em Proceedings of the 2021 CHI Conference on Human Factors in Computing Systems}, pp. 1--12, 2021.

\bibitem{koonchanok2023visual}
R.~Koonchanok, G.~Y. Tawde, G.~R. Narayanasamy, S.~Walimbe, and K.~Reda.
\newblock Visual belief elicitation reduces the incidence of false discovery.
\newblock In {\em Proceedings of the 2023 CHI conference on human factors in computing systems}, pp. 1--17, 2023.

\bibitem{vLLM}
W.~Kwon, Z.~Li, S.~Zhuang, Y.~Sheng, L.~Zheng, C.~H. Yu, J.~E. Gonzalez, H.~Zhang, and I.~Stoica.
\newblock Efficient memory management for large language model serving with pagedattention.
\newblock In {\em Proceedings of the ACM SIGOPS 29th Symposium on Operating Systems Principles}, 2023.

\bibitem{lee2021viral}
C.~Lee, T.~Yang, G.~D. Inchoco, G.~M. Jones, and A.~Satyanarayan.
\newblock Viral visualizations: How coronavirus skeptics use orthodox data practices to promote unorthodox science online.
\newblock In {\em Proceedings of the 2021 CHI Conference on Human Factors in Computing Systems}, CHI '21,  article no. 607,  18 pages. Association for Computing Machinery, New York, NY, USA, 2021. \href{https://doi.org/10.1145/3411764.3445211}
{doi: {{%
10\hspace{.1pt}\discretionary{.}{%
}{.}\hspace{.4pt}1145\discretionary{/}{%
}{/}3411764\hspace{.1pt}\discretionary{.}{%
}{.}\hspace{.4pt}3445211}}}


\bibitem{rag}
P.~Lewis, E.~Perez, A.~Piktus, F.~Petroni, V.~Karpukhin, N.~Goyal, H.~K\"{u}ttler, M.~Lewis, W.-t. Yih, T.~Rockt\"{a}schel, S.~Riedel, and D.~Kiela.
\newblock Retrieval-augmented generation for knowledge-intensive nlp tasks.
\newblock In {\em Proc. NIPS},  article no. 793,  16 pages. Red Hook, NY, USA, 2020.

\bibitem{li2014constructing}
F.~Li and H.~V. Jagadish.
\newblock Constructing an interactive natural language interface for relational databases.
\newblock {\em In Proc. VLDB}, 8(1):73--84, 2014.

\bibitem{li2024kgs}
H.~Li, G.~Appleby, C.~D. Brumar, R.~Chang, and A.~Suh.
\newblock Knowledge graphs in practice: Characterizing their users, challenges, and visualization opportunities.
\newblock {\em IEEE Trans. Vis. Comput. Graph.}, 30(1):584--594, 2024. \href{https://doi.org/10.1109/TVCG.2023.3326904}
{doi: {{%
10\hspace{.1pt}\discretionary{.}{%
}{.}\hspace{.4pt}1109\discretionary{/}{%
}{/}TVCG\hspace{.1pt}\discretionary{.}{%
}{.}\hspace{.4pt}2023\hspace{.1pt}\discretionary{.}{%
}{.}\hspace{.4pt}3326904}}}


\bibitem{li2024linkq}
H.~Li, G.~Appleby, and A.~Suh.
\newblock Linkq: An llm-assisted visual interface for knowledge graph question-answering.
\newblock In {\em 2024 IEEE Visualization and Visual Analytics (VIS)}, pp. 116--120, 2024. \href{https://doi.org/10.1109/VIS55277.2024.00031}
{doi: {{%
10\hspace{.1pt}\discretionary{.}{%
}{.}\hspace{.4pt}1109\discretionary{/}{%
}{/}VIS55277\hspace{.1pt}\discretionary{.}{%
}{.}\hspace{.4pt}2024\hspace{.1pt}\discretionary{.}{%
}{.}\hspace{.4pt}00031}}}


\bibitem{li2024preliminaryroadmapllmsassistants}
H.~Li, G.~Appleby, and A.~Suh.
\newblock A preliminary roadmap for llms as assistants in exploring, analyzing, and visualizing knowledge graphs.
\newblock In {\em IEEE VIS NLVIZ Workshop: Exploring Research Opportunities for Natural Language, Text, and Data Visualization}, 2024.

\bibitem{li2021kg4vis}
H.~Li, Y.~Wang, S.~Zhang, Y.~Song, and H.~Qu.
\newblock Kg4vis: A knowledge graph-based approach for visualization recommendation.
\newblock {\em IEEE Trans. Vis. Comput. Graph.}, 28(01):195--205, 2022. \href{https://doi.org/10.1109/TVCG.2021.3114863}
{doi: {{%
10\hspace{.1pt}\discretionary{.}{%
}{.}\hspace{.4pt}1109\discretionary{/}{%
}{/}TVCG\hspace{.1pt}\discretionary{.}{%
}{.}\hspace{.4pt}2021\hspace{.1pt}\discretionary{.}{%
}{.}\hspace{.4pt}3114863}}}


\bibitem{lipton2018mythos}
Z.~C. Lipton.
\newblock The mythos of model interpretability: In machine learning, the concept of interpretability is both important and slippery.
\newblock {\em Queue}, 16(3):31--57, 2018.

\bibitem{lisnic2023misleading}
M.~Lisnic, C.~Polychronis, A.~Lex, and M.~Kogan.
\newblock Misleading beyond visual tricks: How people actually lie with charts.
\newblock In {\em Proceedings of the 2023 CHI Conference on Human Factors in Computing Systems}, CHI '23,  article no. 817,  21 pages. Association for Computing Machinery, New York, NY, USA, 2023. \href{https://doi.org/10.1145/3544548.3580910}
{doi: {{%
10\hspace{.1pt}\discretionary{.}{%
}{.}\hspace{.4pt}1145\discretionary{/}{%
}{/}3544548\hspace{.1pt}\discretionary{.}{%
}{.}\hspace{.4pt}3580910}}}


\bibitem{lissandrini2022knowledge}
M.~Lissandrini, D.~Mottin, K.~Hose, and T.~B. Pedersen.
\newblock Knowledge graph exploration systems: are we lost?
\newblock In {\em CIDR}, vol.~22, pp. 10--13, 2022.

\bibitem{lissandrini2020graph}
M.~Lissandrini, D.~Mottin, T.~Palpanas, and Y.~Velegrakis.
\newblock Graph-query suggestions for knowledge graph exploration.
\newblock In {\em Proc. ACM WWW}, pp. 2549--2555, 2020.

\bibitem{knowledgeInjecitonToCounter}
A.~Martino, M.~Iannelli, and C.~Truong.
\newblock Knowledge injection to counter large language model (llm) hallucination.
\newblock In C.~Pesquita, H.~Skaf-Molli, V.~Efthymiou, S.~Kirrane, A.~Ngonga, D.~Collarana, R.~Cerqueira, M.~Alam, C.~Trojahn, and S.~Hertling, eds., {\em The Semantic Web: ESWC 2023 Satellite Events}, pp. 182--185. Springer Nature Switzerland, Cham, 2023.

\bibitem{mckinley2025trustworthy}
O.~McKinley, S.~Pandey, and A.~Ottley.
\newblock Trustworthy by design: The viewer's perspective on trust in data visualization.
\newblock {\em arXiv preprint arXiv:2503.10892}, 2025.

\bibitem{miller2013graph}
J.~J. Miller.
\newblock Graph database applications and concepts with {Neo4J}.
\newblock {\em Proc. SAIS}, 2324(36), 2013.

\bibitem{mitra2022facilitating}
R.~Mitra, A.~Narechania, A.~Endert, and J.~Stasko.
\newblock Facilitating conversational interaction in natural language interfaces for visualization.
\newblock In {\em Proc. VIS}, pp. 6--10, 2022. \href{https://doi.org/10.1109/VIS54862.2022.00010}
{doi: {{%
10\hspace{.1pt}\discretionary{.}{%
}{.}\hspace{.4pt}1109\discretionary{/}{%
}{/}VIS54862\hspace{.1pt}\discretionary{.}{%
}{.}\hspace{.4pt}2022\hspace{.1pt}\discretionary{.}{%
}{.}\hspace{.4pt}00010}}}


\bibitem{Narechania2020NL4DVAT}
A.~{Narechania}, A.~{Srinivasan}, and J.~{Stasko}.
\newblock {NL4DV}: A {Toolkit} for generating {Analytic Specifications} for {Data Visualization} from {Natural Language} queries.
\newblock {\em IEEE Trans. Vis. Comput. Graph.}, 2020. \href{https://doi.org/10.1109/TVCG.2020.3030378}
{doi: {{%
10\hspace{.1pt}\discretionary{.}{%
}{.}\hspace{.4pt}1109\discretionary{/}{%
}{/}TVCG\hspace{.1pt}\discretionary{.}{%
}{.}\hspace{.4pt}2020\hspace{.1pt}\discretionary{.}{%
}{.}\hspace{.4pt}3030378}}}


\bibitem{ngonga2013sorry}
A.-C. Ngonga~Ngomo, L.~B\"{u}hmann, C.~Unger, J.~Lehmann, and D.~Gerber.
\newblock Sorry, i don't speak sparql: translating sparql queries into natural language.
\newblock In {\em Proc. ACM WWW},  12 pages, p. 977–988. ACM, New York, NY, USA, 2013. \href{https://doi.org/10.1145/2488388.2488473}
{doi: {{%
10\hspace{.1pt}\discretionary{.}{%
}{.}\hspace{.4pt}1145\discretionary{/}{%
}{/}2488388\hspace{.1pt}\discretionary{.}{%
}{.}\hspace{.4pt}2488473}}}


\bibitem{north2006toward}
C.~North.
\newblock Toward measuring visualization insight.
\newblock {\em IEEE computer graphics and applications}, 26(3):6--9, 2006.

\bibitem{2024_unifying_llms_and_kgs}
S.~Pan, L.~Luo, Y.~Wang, C.~Chen, J.~Wang, and X.~Wu.
\newblock Unifying large language models and knowledge graphs: A roadmap.
\newblock {\em IEEE Trans. Knowl. Data Eng.}, pp. 1--20, 2024. \href{https://doi.org/10.1109/TKDE.2024.3352100}
{doi: {{%
10\hspace{.1pt}\discretionary{.}{%
}{.}\hspace{.4pt}1109\discretionary{/}{%
}{/}TKDE\hspace{.1pt}\discretionary{.}{%
}{.}\hspace{.4pt}2024\hspace{.1pt}\discretionary{.}{%
}{.}\hspace{.4pt}3352100}}}


\bibitem{petroni2019language}
F.~Petroni, T.~Rockt{\"a}schel, S.~Riedel, P.~Lewis, A.~Bakhtin, Y.~Wu, and A.~Miller.
\newblock Language models as knowledge bases?
\newblock In {\em Proc. EMNLP/IJCNLP}, pp. 2463--2473. ACL, Hong Kong, 2019. \href{https://doi.org/10.18653/v1/D19-1250}
{doi: {{%
10\hspace{.1pt}\discretionary{.}{%
}{.}\hspace{.4pt}18653\discretionary{/}{%
}{/}v1\discretionary{/}{%
}{/}D19\discretionary{%
}{-}{-}1250}}}


\bibitem{rangel2024sparql}
J.~C. Rangel, T.~M. de~Farias, A.~C. Sima, and N.~Kobayashi.
\newblock Sparql generation: an analysis on fine-tuning openllama for question answering over a life science knowledge graph.
\newblock {\em arXiv preprint arXiv:2402.04627}, 2024.

\bibitem{rawte2023survey}
V.~Rawte, A.~Sheth, and A.~Das.
\newblock A survey of hallucination in large foundation models.
\newblock {\em arXiv preprint arXiv:2309.05922}, 2023.

\bibitem{rillig2023risks}
M.~C. Rillig, M.~{\AA}gerstrand, M.~Bi, K.~A. Gould, and U.~Sauerland.
\newblock Risks and benefits of large language models for the environment.
\newblock {\em Environmental Science \& Technology}, 57(9):3464--3466, 2023.

\bibitem{rogers2012hci}
Y.~Rogers.
\newblock {\em HCI theory: classical, modern, and contemporary}, vol.~14.
\newblock Morgan \& Claypool Publishers, 2012.

\bibitem{selmdmair2012dsm}
M.~Sedlmair, M.~Meyer, and T.~Munzner.
\newblock Design study methodology: Reflections from the trenches and the stacks.
\newblock {\em IEEE Trans. Vis. Comput. Graph.}, 18(12):2431--2440, 2012. \href{https://doi.org/10.1109/TVCG.2012.213}
{doi: {{%
10\hspace{.1pt}\discretionary{.}{%
}{.}\hspace{.4pt}1109\discretionary{/}{%
}{/}TVCG\hspace{.1pt}\discretionary{.}{%
}{.}\hspace{.4pt}2012\hspace{.1pt}\discretionary{.}{%
}{.}\hspace{.4pt}213}}}


\bibitem{sen2022mintaka}
P.~Sen, A.~F. Aji, and A.~Saffari.
\newblock Mintaka: A complex, natural, and multilingual dataset for end-to-end question answering.
\newblock {\em arXiv preprint arXiv:2210.01613}, 2022.

\bibitem{setlur2016eviza}
V.~Setlur, S.~E. Battersby, M.~Tory, R.~Gossweiler, and A.~X. Chang.
\newblock Eviza: A natural language interface for visual analysis.
\newblock In {\em In Proc. ACM UIST}, pp. 365--377, 2016.

\bibitem{srinivasan2021snowy}
A.~Srinivasan and V.~Setlur.
\newblock Snowy: Recommending utterances for conversational visual analysis.
\newblock In {\em ACM UIST}, pp. 864--880, 2021.

\bibitem{suh2024luminate}
S.~Suh, M.~Chen, B.~Min, T.~J.-J. Li, and H.~Xia.
\newblock Luminate: Structured generation and exploration of design space with large language models for human-ai co-creation.
\newblock In {\em Proceedings of the CHI Conference on Human Factors in Computing Systems}, pp. 1--26, 2024.

\bibitem{sultanum2024data}
N.~Sultanum, D.~Bromley, and M.~Correll.
\newblock Data guards: Challenges and solutions for fostering trust in data.
\newblock In {\em 2024 IEEE Visualization and Visual Analytics (VIS)}, pp. 56--60. IEEE, 2024.

\bibitem{sultanum2023datatales}
N.~Sultanum and A.~Srinivasan.
\newblock Datatales: Investigating the use of large language models for authoring data-driven articles.
\newblock In {\em Proc. VIS}, pp. 231--235, 2023.

\bibitem{tian2024chartgpt}
Y.~Tian, W.~Cui, D.~Deng, X.~Yi, Y.~Yang, H.~Zhang, and Y.~Wu.
\newblock Chartgpt: Leveraging llms to generate charts from abstract natural language.
\newblock {\em IEEE Transactions on Visualization and Computer Graphics}, 2024.

\bibitem{vargas2019rdf}
H.~Vargas, C.~Buil-Aranda, A.~Hogan, and C.~L{\'o}pez.
\newblock Rdf explorer: A visual sparql query builder.
\newblock In {\em The Semantic Web--ISWC 2019: 18th International Semantic Web Conference, Auckland, New Zealand, October 26--30, 2019, Proceedings, Part I 18}, pp. 647--663. Springer, 2019.

\bibitem{w3corg}
W.~W. W.~C. (W3C).
\newblock {RDF Primer}.
\newblock \url{https://www.w3.org/TR/rdf11-primer/}.
\newblock Accessed: 2024-08-26.

\bibitem{wang2024promptcharm}
Z.~Wang, Y.~Huang, D.~Song, L.~Ma, and T.~Zhang.
\newblock Promptcharm: Text-to-image generation through multi-modal prompting and refinement.
\newblock In {\em Proceedings of the CHI Conference on Human Factors in Computing Systems}, pp. 1--21, 2024.

\bibitem{wei2020combining}
H.~Wei.
\newblock Combining knowledge graphs, quickly and accurately.
\newblock {\em Amazon Science}, 2020.

\bibitem{wei2022chain}
J.~Wei, X.~Wang, D.~Schuurmans, M.~Bosma, F.~Xia, E.~Chi, Q.~V. Le, D.~Zhou, et~al.
\newblock Chain-of-thought prompting elicits reasoning in large language models.
\newblock {\em Advances in neural information processing systems}, 35:24824--24837, 2022.

\bibitem{wen2024mindmap}
Y.~Wen, Z.~Wang, and J.~Sun.
\newblock Mindmap: Knowledge graph prompting sparks graph of thoughts in large language models.
\newblock In {\em Proc. Association for Computational Linguistics}, 2024.

\bibitem{wikidata2024stats}
Wikipedia.
\newblock {Wikidata Statistics}.
\newblock \url{https://www.wikidata.org/wiki/Wikidata:Statistics}.
\newblock Accessed: 2025-03-18.

\bibitem{yang2023llm}
S.~Yang, M.~Teng, X.~Dong, and F.~Bo.
\newblock Llm-based sparql generation with selected schema from large scale knowledge base.
\newblock In {\em China Conference on Knowledge Graph and Semantic Computing}, pp. 304--316. Springer, 2023.

\bibitem{zhang2023dissecting}
L.~Zhang, X.~Liu, Z.~Li, X.~Pan, P.~Dong, R.~Fan, R.~Guo, X.~Wang, Q.~Luo, S.~Shi, et~al.
\newblock Dissecting the runtime performance of the training, fine-tuning, and inference of large language models.
\newblock {\em arXiv preprint arXiv:2311.03687}, 2023.

\end{thebibliography}

\end{document}